\documentclass[10pt,journal,compsoc]{IEEEtran}

\ifCLASSOPTIONcompsoc
  \usepackage[nocompress]{cite}
\else
  \usepackage{cite}
\fi

\ifCLASSINFOpdf

\else

\fi

\usepackage[colorlinks,
linkcolor=red,
anchorcolor=green,
citecolor=blue
]{hyperref}
\usepackage{url}            
\usepackage{booktabs}       
\usepackage{amsfonts}       
\usepackage{nicefrac}       
\usepackage{microtype}      
\usepackage{xcolor}         

\usepackage{amsmath}
\usepackage{graphicx}
\usepackage{algorithm}
\usepackage{algorithmic}
\usepackage{makecell}

\begin{document}
\title{Ensemble Multi-Quantiles: Adaptively Flexible Distribution Prediction for Uncertainty Quantification}


\author{Xing Yan, 
        Yonghua Su, 
        and Wenxuan Ma
\IEEEcompsocitemizethanks{\IEEEcompsocthanksitem All authors are with the Institute of Statistics and Big Data, Renmin University of China, 
Beijing 100872, China.\protect\\
E-mail: \{xingyan,2019000154,mawenxuan\}@ruc.edu.cn.
}
}


\markboth{Ensemble Multi-Quantiles: Adaptively Flexible Distribution Prediction for Uncertainty Quantification}%
{Shell \MakeLowercase{\textit{et al.}}: Bare Demo of IEEEtran.cls for Computer Society Journals}

\IEEEtitleabstractindextext{%
\begin{abstract}
We propose a novel, succinct, and effective approach for distribution prediction to quantify uncertainty in machine learning. It incorporates adaptively flexible distribution prediction of $\mathbb{P}(\mathbf{y}|\mathbf{X}=x)$ in regression tasks. This conditional distribution's quantiles of probability levels spreading the interval $(0,1)$ are boosted by additive models which are designed by us with intuitions and interpretability. We seek an adaptive balance between the structural integrity and the flexibility for $\mathbb{P}(\mathbf{y}|\mathbf{X}=x)$, while Gaussian assumption results in a lack of flexibility for real data and highly flexible approaches (e.g., estimating the quantiles separately without a distribution structure) inevitably have drawbacks and may not lead to good generalization. This ensemble multi-quantiles approach called EMQ proposed by us is totally data-driven, and can gradually depart from Gaussian and discover the optimal conditional distribution in the boosting. On extensive regression tasks from UCI datasets, we show that EMQ achieves state-of-the-art performance comparing to many recent uncertainty quantification methods. Visualization results further illustrate the necessity and the merits of such an ensemble model.
\end{abstract}

\begin{IEEEkeywords}
Uncertainty Quantification, Distribution Prediction, Adaptive Flexibility, Non-Gaussianity, Ensemble Multi-Quantiles
\end{IEEEkeywords}}

\maketitle

\IEEEdisplaynontitleabstractindextext
\IEEEpeerreviewmaketitle

\IEEEraisesectionheading{\section{Introduction}\label{sec:introduction}}
\IEEEPARstart{M}{achine} learning algorithms have become very powerful in recent years due to the successes of deep neural networks. Advanced deep learning models have achieved state-of-the-art performance in many tasks under the evaluation criteria of prediction accuracy \cite{lecun2015deep}. However, people criticize deep learning for not only its opaque in making predictions but also its over-confidence on the predictions \cite{guo2017calibration,rahaman2021uncertainty}.
Over-confidence may occur when deep learning has a wrong estimation of the uncertainty.
We thereby need giving a good measurement of the uncertainty level for the prediction for each input of the model, e.g., giving a standard deviation or a prediction interval (that the label will fall into with a fixed probability) accompanied with the prediction value in a regression task.

Uncertainty quantification lies in a crucial position in machine learning, especially in real applications of deep learning when people must make decisions relying on it. A wrong estimate of uncertainty will make decisions highly risky. Good uncertainty estimation quantifies when we can trust the model's predictions. For example, in disease detection \cite{begoli2019need}, the model can defer to human experts when it believes the uncertainty is high. 
An autopilot can transfer the authority to the human operator when the prediction is risky.
Besides, good uncertainty quantification can benefit us in detecting and dealing with out-of-distribution samples \cite{lakshminarayanan2017simple}, whose uncertainty should be quantified to be very high.
Another representative case is in financial risk management, decision-makers need to set a new margin rate or to take a hedging action if he/she believes the market uncertainty is high, or, the risk measure defined by some distributional characteristic of the market variable (e.g., the change of S\&P 500 index in a future period) is high. Such distributional characteristics include volatility, kurtosis, Value-at-Risk, expected shortfall, etc., all of which are important and commonly-used risk measures in financial markets \cite{mcneil2015quantitative,duffie1997overview,acerbi2002coherence}.

It has been documented in many existing works \cite{gal2016dropout,lakshminarayanan2017simple,maddox2019simple,rahaman2021uncertainty} that deep learning may produce poor uncertainty estimation. This is partially because there is a lack of ground truth for the uncertainty. Thus, solving this important problem is facing enormous challenges.
For classification tasks where the softmax technique is used, it is straightforward to take the prediction score of the class as uncertainty estimation. But researchers rejected this practice, e.g., in \cite{gal2016dropout}, \cite{sensoy2018evidential}. Although the uncertainty quantification in classification tasks needs much attention, in this paper, we focus on the regression problem whose uncertainty estimation is challenging too. In regression, (aleatoric) uncertainty quantification can be formulated as the problem of estimating some distributional characteristics of the conditional distribution $\mathbb{P}(\mathbf{y}|\mathbf{X}=x)$, where $\mathbf{X}$ is the feature vector and $\mathbf{y}$ is the continuous  label.

Recently, a variety of methods have been developed for uncertainty quantification for neural networks.
A direct approach is to assume a Gaussian distribution for $\mathbb{P}(\mathbf{y}|\mathbf{X}=x)$ and try to estimate its standard deviation as a function of $x$, using a neural network trained with maximum likelihood and some special scheme, see example works such as \cite{lakshminarayanan2017simple,kuleshov2018accurate,skafte2019reliable,zhao2020individual}. This simple idea has been proven very successful when good training and the ensemble are deployed.
Besides, Bayesian methods can naturally return the density function of $\mathbb{P}(\mathbf{y}|\mathbf{X}=x)$, such as  in \cite{hernandez2015probabilistic,gal2016dropout,maddox2019simple}. 
Bayesian methods are sometimes hard to implement and computationally expensive \cite{lakshminarayanan2017simple}, and they require the prior distribution to be correctly specified (but usually, priors of convenience are assumed).
Conformal prediction \cite{shafer2008tutorial,balasubramanian2014conformal} is also a popular framework which separates a calibration set from the training set and quantifies uncertainty with a conformity score. Many variants \cite{papadopoulos2011regression,lei2013distribution,romano2019conformalized} were proposed to make conformal prediction more flexible and adaptive to heteroscedasticity.
Some recent approaches estimate the conditional distribution by designing some novel loss function between the assumed $\mathbb{P}(\mathbf{y}|\mathbf{X}=x)$ and the observed data, such as in \cite{cui2020calibrated,zhou2021estimating}.
The ensemble over multiple Gaussians \cite{wenzel2020hyperparameter} or prediction intervals \cite{pearce2018high} was proposed to obtain better uncertainty estimation, with the parallel ensemble manner (not the boosting-like in this paper).
To summarize, there are many different types of methods for uncertainty quantification in machine learning.

\subsection{Motivations}\label{sec:motivations}

There is still another class of methods that quantify uncertainty by estimating quantiles \cite{song2019distribution,tagasovska2019single,romano2019conformalized,chung2021beyond,feldman2021improving} to construct prediction intervals, which is also the main focus of this paper. Most existing works employ the quantile regression loss from \cite{koenker1978regression,koenker2001quantile} to do the model training. They generally will produce a prediction interval $[q_{\tau}(x),q_{1-\tau}(x)]$ formed by the $\tau$-quantile and $(1-\tau)$-quantile of $\mathbb{P}(\mathbf{y}|\mathbf{X}=x)$. However, we focus on the situation where many $[q_{\tau}(x),q_{1-\tau}(x)]$ need to be predicted for many different $\tau$ (for quantifying uncertainty with different confidences). From another perspective, those quantiles can form an approximation of $\mathbb{P}(\mathbf{y}|\mathbf{X}=x)$. The focus now becomes the conditional distribution prediction.

In this case, if no distribution structure is assumed, it is equivalent to assuming an extremely flexible distribution for $\mathbb{P}(\mathbf{y}|\mathbf{X}=x)$. It corresponds to the non-parametric approach in statistics context.
The opposite extreme is assuming a commonly-used and well-understood distribution structure such as Gaussian and then predicting its parameters. We argue that both the two extremes are potentially harmful for better uncertainty estimation.

In the former approach, the resulting implied density function (can be obtained from the quantiles predicted) will display a chaotic function graph with no distribution shape, even if we let the training dataset be large. 
One can refer to Fig. \ref{fig:other_density} for a clear illustration. Some methods developed a neural-network representation for $\mathbb{P}(\mathbf{y}|\mathbf{X}=x)$ (e.g., modeling $q_{\tau}(x)$ as a neural network taking $x$ and $\tau\in(0,1)$ as inputs), which also incorporate great flexibility in distribution prediction. 
We argue that too much flexibility does not necessarily lead to performance improvement because: i) the training will become more difficult; and ii) it may be easier to overfit; and iii) it will lead to strange-shaped or even invalid density function with no interpretability. Unnecessary flexibility may be harmful, and has been added into the modeling of $\mathbb{P}(\mathbf{y}|\mathbf{X}=x)$ in some existing methods.
We will show that seeking a balance between the flexibility and the distribution structure will produce substantially better distribution prediction. Actually, by proposing a novel ensemble multi-quantiles (EMQ) method, we successfully obtain adaptively flexible distribution prediction. It can find the balance adaptively and thus approximate $\mathbb{P}(\mathbf{y}|\mathbf{X}=x)$ well, as we will show in the rest of the paper.

Oppositely, in the other approach, assuming a succinct distribution structure for $\mathbb{P}(\mathbf{y}|\mathbf{X}=x)$ has obvious shortcomings. It will suffer from the lack of flexibility, and the model performance will heavily rely on the correctness of the choice of distribution structure.
For convenience, usually Gaussian is chosen. However, conditional distributions in real data may exhibit multi-modality, asymmetry, and heavy tails, and may even be heterogeneous within a dataset. 
The mixture of Gaussians in Bayesian approaches or model ensembles does not alleviate this kind of deficiency substantially.
Other distribution structures such as $t$-distribution, skew-$t$, and Generalized beta, may have complicated mathematical expressions for their density functions, increasing the difficulty of training a model with them with maximum likelihood loss because the gradient with respect to the distribution parameters will be difficult to obtain. Additionally, the choice among them relies on specific domain knowledge.


Another motivation of this paper is to solve the quantile-crossing issue when estimating multi-quantiles of $\mathbb{P}(\mathbf{y}|\mathbf{X}=x)$. Although there have been many methods proposed in statistics literature \cite{he1997quantile,takeuchi2006nonparametric,liu2009stepwise,bondell2010noncrossing,chernozhukov2010quantile}, they scarcely solve the unnecessary flexibility issue we mention above, yet may cause other problems such as the computational burden/difficulty if they are constraint optimization-based. 
Besides constraint optimization, post-processing or re-sorting is another imperfect solution to the quantile-crossing issue.
We believe that more natural solutions without additional effort to deal with such an issue are absolutely necessary. Actually, the proposed EMQ naturally overcomes this issue with nice designs. Our solution is a built-in solution without additional effort such as constraint optimization or post-processing, as shown in the rest of the paper.

\subsection{Our Contributions}

In this paper, we tackle the problem of conditional distribution prediction in regression tasks through proposing an ensemble learning approach named EMQ. Multi-quantiles are boosted by additive models with intuitive designs, and can be used for uncertainty quantification with excellent performance. To summarize, our contributions are as follows:

\textbf{i)} We propose a novel, succinct, and effective approach for multi-quantiles ensemble learning, in which a strong learner aims at learning a Gaussian distribution for $\mathbb{P}(\mathbf{y}|\mathbf{X}=x)$
and the consequent weak learners are used for tuning or improving the quantiles. The design of the whole ensemble approach is intuitive and interpretable, and aims at adaptively seeking a balance between the distribution structure (such as Gaussian) and the flexibility. 

\textbf{ii)} The quantile-crossing issue is overcome naturally in our approach without additional effort.

\textbf{iii)} In the experiments, the proposed EMQ achieves start-of-the-art performance over a large range of datasets under the criteria of calibration and sharpness, compared to many recent uncertainty quantification methods including Gaussian assumption-based, Bayesian methods, quantile regression-based, and traditional tree models.

\textbf{iv)} Through the experiments, we find that EMQ can indeed do adaptively flexible distribution prediction, especially when we adopt the adaptive $T_{ada}$ strategy. The ensemble steps in EMQ work correctly as we except. In addition, the variant EMQW adopts the weighted quantile loss and achieves superior performance on the tail sides.

\textbf{v)} Empirically, we find that some datasets have Gaussian-distributed $\mathbb{P}(\mathbf{y}|\mathbf{X}=x)$ while others have non-Gaussian ones with significantly different density shapes such as sharp peak, asymmetry, long tail, multi-modality, etc., all of which are successfully captured by our approach. Even in one dataset, this variety may occur, which verifies the complexity of real data and the necessity and merits of our approach.

Code and data are released in the link\footnote{\href{https://github.com/xingyan-fml/emq}{https://github.com/xingyan-fml/emq}}.

\section{Related Works}

We summarize some recent popular works that are directly related to our research, which will be the competing methods in our experiments. They can be divided into three categories: variance networks, Bayesian, and quantile regression-based.

\textbf{Variance networks.} It is straightforward to let the neural network output a variance to represent the uncertainty. Heteroskedasticity Neural Network (HNN) is such a method that outputs both the mean (the prediction) and the variance. Gaussian maximum likelihood loss is used to learn the network parameters. \cite{skafte2019reliable} suggested to separately and alternately train a mean network and a variance network, for better variance estimation. We will follow this suggestion in our implementation of HNN in the experiments. Deep Ensemble is the method who will train multiple HNNs with different initializations and aggregate the means and variances obtained (see \cite{lakshminarayanan2017simple}). Despite being popular, these two methods are Gaussian assumption-based and are not suitable for more complicated non-Gaussian cases.

\textbf{Bayesian.} Bayesian methods compute the predictive posterior distribution as the uncertainty estimation by weighting the likelihood with the posterior of network parameters. MC Dropout \cite{gal2016dropout} uses dropout as a Bayesian approximation to overcome the computational challenge. Concrete Dropout \cite{gal2017concrete} is a dropout variant which allows the dropout probability to be tuned continuously using gradient methods. As stated in original papers, these two Bayesian methods are popular for large computer vision models. Nevertheless, they do not show superior performance on tabular data or other structured data. In addition, similar to variance networks, they only output a variance as the uncertainty estimation, without more details on the conditional distribution.

\textbf{Quantile regression-based.} Quantile regression has a long history in statistics literature \cite{koenker1978regression,koenker2001quantile,koenker2006quantile}. The loss function in quantile regression is $L_{\tau} (y,q) =  (y-q)(\tau-\mathbb{I}_{\{y<q\}})$, where $y$ is the observed label, $q$ is the $\tau$-quantile predicted by a model given $x$, and $\tau$ is a specified probability level. If a model can output multiple quantiles of different levels, the losses for different $\tau$ can be summed and then minimized. We call this approach vanilla quantile regression (Vanilla QR). 
\cite{tagasovska2019single} proposed simultaneous quantile regression (SQR), a neural network model which takes both the feature vector $x$ and the level $\tau$ as inputs, and outputs the $\tau$-quantile, and minimizes the above pinball loss with $\tau\sim \text{Uniform}(0,1)$.
An alternative to the pinball loss is the interval score proposed in \cite{gneiting2007strictly}. Let $(l, u )$ be the $(1-\tau)$ centered prediction interval, where $l$ and $u$ stand for the $\frac{\tau}{2}$-quantile and $(1-\frac{\tau}{2})$-quantile we are looking for. This yields the interval score:
\begin{equation}
S_{\tau}(l, u ; y)=(u-l)+\frac{2}{\tau}(l-y) \mathbb{I}_{\{y<l\}}+\frac{2}{\tau}(y-u) \mathbb{I}_{\{y>u\}}. \label{eqn:interval_score}
\end{equation}
Besides, \cite{chung2021beyond} proposed a combined loss considering both calibration and sharpness terms and firstly simultaneously optimized the interval score. \cite{feldman2021improving} proposed the pinball loss with an orthogonality regularization by considering the independence between the interval length and the violation.

\textbf{Other methods of probabilistic forecasting.} \cite{gasthaus2019probabilistic} adopted a non-parametric method, the linear spline, or a piecewise-linear function to represent a conditional quantile function. The parameters of the spline are the output of a recurrent neural network with the features as its input. It may need many pieces in such a piecewise-linear function to approximate a smooth quantile function well, e.g., the quantile function of a simple Gaussian.
In financial econometrics literature, \cite{theodossiou1998financial,bali2008role} proposed a skewed generalized $t$-distribution, which has a highly complicated mathematical form and of course is computationally unfriendly, especially in the conditional setting where the gradient of the loss will be difficult to obtain. \cite{yan2019cross} adopted a new parametric quantile function which transforms the Gaussian. It has four parameters that control the location, scale, right tail, and left tail respectively. The new quantile function has an S-shaped Q-Q plot against Gaussian, thus with heavier tails, but with limited flexibility.

\subsection{Aleatoric and Epistemic Uncertainty}

Aleatoric uncertainty \cite{der2009aleatory} refers to the randomness of the outcome, or, the inherently stochastic and irreducible random effects. It does not vanish as the sample size goes to infinity.
Epistemic uncertainty \cite{hullermeier2021aleatoric}, or model uncertainty, refers to the uncertainty caused by a lack of knowledge. It is the uncertainty due to model limitations.
Generally, Bayesian and ensemble methods can estimate both epistemic and aleatoric uncertainty \cite{skafte2019reliable}. The variance networks and quantile-based methods focus on the aleatoric uncertainty. In this paper, we focus on the aleatoric uncertainty only too, as what we propose is intrinsically a quantile-based approach. 
Both types of uncertainty are important and challenging to estimate.
For more about epistemic uncertainty estimation, please refer to \cite{tagasovska2019single,lahlou2021deup}, etc.

\section{Methodology}

In this section, we elaborate the proposed ensemble multi-quantiles (EMQ) approach for uncertainty quantification. Throughout the paper, we use $\mathbf{X}$ and $\mathbf{y}$ to denote the random variables of the features and the continuous label. Given the feature vector $\mathbf{X}=x$, the goal is to predict the conditional $\tau_k$-quantile of the variable $\mathbf{y}$, denoted as $q_k(x)$, where $k=1,\dots,K$ and $0<\tau_1<\cdots<\tau_K<1$. 
Here $\tau_1,\dots,\tau_K$ are probability levels specified by us that densely distribute in the interval $(0,1)$, such as $0.01,0.02,\dots,0.98,0.99$, denoted as the set $\Pi=\{\tau_k\}_{k=1}^K$.
These $K$ conditional quantiles form an approximation of the conditional distribution $\mathbb{P}(\mathbf{y}|\mathbf{X}=x)$.
The learning will be conducted given a dataset $\{(x_i,y_i)\}_{i\in D_{tr}}$.
For concision, we sometimes drop the letter $x$ and simplify the notation $q_k(x)$ to $q_k$, and suppose $x$ is given and remains the same in this situation.

\subsection{The Framework}

Our general idea is to iteratively update the quantile predictions in multiple steps in a boosting manner, i.e., updating $q^t_1,\dots,q^t_K$ as the step $t=0,1,\dots,T$ with additive models and taking the final-step predictions $q^T_1,\dots,q^T_K$ as the desired ones. To be more specific, denoting $Q^t(x) = [q^t_1(x),\dots,q^t_K(x)]^\top$, we propose a new ensemble framework:
\begin{align}
Q^0(x) & = G_0\left(F_0(x;\Theta_0)\right), \label{eqn:Q^0}\\
Q^t(x) & = Q^{t-1}(x)+G\left(F_t(x;\Theta_t),Q^{t-1}(x)\right), ~~ t=1,\dots,T. \label{eqn:Q^t}
\end{align}
$F_0$ and $F_t$ are neural network models with learnable parameters $\Theta_0$ and $\Theta_t$. $G_0$ and $G$ are fixed transformations that convert the neural network outputs to the quantile predictions, as we will introduce later. 

Iteratively, at every step $t=0,\dots,T$, a learning objective is set by us to train the model $F_t$ with $F_0,\dots,F_{t-1}$ fixed:
\begin{equation}
\min_{\Theta_t} \mathbb{E}_{\mathbb{P}(\mathbf{X},\mathbf{y})} \left[ L(\mathbf{y},Q^t(\mathbf{X})) \right].
\end{equation}
$L$ is a proper loss function between $y$ and $Q^t(x)$. 
Given the probability level set $\Pi=\{\tau_k\}_{k=1}^{K}$, the loss $L$ is chosen as the sum of quantile regression losses:
\begin{align}
L(y,Q^t(x)) = \sum_{k=1}^K L_{\tau_k} (y,q^t_k(x)), \label{eqn:loss_L}\\
L_{\tau} (y,q) =  (y-q)(\tau-\mathbb{I}_{\{y<q\}}). \label{eqn:loss_L_tau}
\end{align}
Given the training dataset, the empirical version of the learning objective that we should optimize is
\begin{equation}
\min_{\Theta_t}  \frac{1}{|D_{tr}|} \sum_{i\in D_{tr}} L(y_i,Q^t(x_i)) + \lambda_t R(\Theta_t). \label{eqn:final_learning_goal}
\end{equation}
$R(\Theta_t)$ is a regularization term and $\lambda_t$ is a trade-off hyper-parameter. The above framework is general and needs to be elaborated further. In the following, we carefully design every component of it, particularly the choices for $F_0,G_0,F_t,G$ and the dealing of regularization term $\lambda_t R(\Theta_t)$.

\subsection{Intuitions}

\textbf{The initial step $\mathbf{t=0}$.} In traditional boosting methods, weak learners are boosted iteratively to form a strong learner. The weak learner at $t=0$ may be a constant (such as in GBDT) or have no difference to those at $t>0$. Here we face a different situation.
Because we are motivated by i) the necessary adaptive flexibility of the distribution that departs from Gaussian, and ii) the fact that assuming Gaussian distribution can yield acceptable uncertainty estimation sometimes. Thus, starting from Gaussian will be a natural and reasonable idea. Consequently, we aim to let the quantile predictions $Q^0(x)$ at $t=0$ be as good as possible under the Gaussian assumption. So, $F_0$ and $G_0$ will be designed to align with this goal. $F_0$ cannot be treated as a weak learner now.
This will make the following learning at $t>0$ relatively easy.

\textbf{The ensemble steps $\mathbf{t>0}$.} After learning a Gaussian-distributed $\mathbb{P}(\mathbf{y}|\mathbf{X}=x)$, imposing consecutive moderate modifications or tuning on the distribution is a natural idea, and is also consistent with the motivations of this paper. So, when $t>0$, we set $F_t$ as a relatively weak learner that aims to moderately update the quantile predictions $Q^{t-1}(x)$ to better ones $Q^{t}(x)$. This updating should not be totally free or non-parametric, because preserving partially the old distribution is necessary for the purpose of adaptive flexibility. $F_t$ and $G$ will be designed accordingly. The transformer $G$ which is a crucial part in the framework builds a bridge between the weak learner $F_t$ and the quantile differences $Q^{t}(x)-Q^{t-1}(x)$ in two adjacent steps, as shown later.

\textbf{The regularization term.} Traditional machine learning models may adopt explicit regularization terms added into the learning objective, to decrease the model complexity and reduce the risk of over-fitting. 
For deep neural networks, the regularization terms can be in implicit forms, such as batch normalization, dropout, and early stopping, all of which are popular strategies. In our framework, both $F_0$ and $F_t$ are neural networks. Thus, we adopt the following regularization strategies instead of the explicit term $\lambda_t R(\Theta_t)$: i) restrict the number of layers and the hidden layer sizes in $F_t$ when $t>0$; ii) restrict the output dimension of $F_t$ for $t\ge 0$ (with the assistance of non-learnable $G_0$ or $G$ to map the output to the quantiles); and iii) adopt those strategies mentioned above such as early stopping. Finally and most importantly, to further avoid over-fitting and to seek the balance between the flexibility and the distribution structure, we use a regularization strategy of adaptively deciding the number of ensemble steps $T$, called the adaptive $T$ strategy.

\textbf{The monotonic property of quantiles.} As we mentioned in Section \ref{sec:motivations}, a natural and built-in solution to the quantile-crossing issue is required in our approach. Because $Q^0(x)$ obeys a Gaussian distribution, the monotonic property is of course satisfied. We suppose $Q^{t-1}(x)$ satisfies the monotonic property, i.e., $q^{t-1}_1<\cdots<q^{t-1}_K$. As long as we have an appropriate transformer $G$ whose output is the difference $Q^t(x)-Q^{t-1}(x)$, we can yield monotonically increasing $Q^t(x)$: $q^{t}_1<\cdots<q^{t}_K$. Thereby, we can solve the quantile-crossing issue naturally without additional effort. Again we see $G$ is a crucial part in our framework. We will design an appropriate $G$ that meets this goal.

To summarize, starting from Gaussian at $t=0$, we gradually and moderately update $q^t_1,\dots,q^t_K$ to be better, and meanwhile ensure that $q^t_1,\dots,q^t_K$ are strictly increasing with respect to $k$.
The relatively strong learner $F_0$ and the weak learners $F_t$, $t>0$ will be trained. Appropriate $G_0$ and $G$ will be designed and useful regularization strategies will be adopted, including the adaptive $T$ strategy.
The resulting final-step predictions $q^T_1,\dots,q^T_K$ are monotonically increasing and can form a good approximation of the conditional distribution $\mathbb{P}(\mathbf{y}|\mathbf{X}=x)$.
We will elaborate all of these in the following.

\subsection{The Initial Step}

At $t=0$, Gaussian assumption gives $\mathbb{P}(\mathbf{y}|\mathbf{X}=x) \sim \mathcal{N}(\mu(x),\sigma(x)^2)$. We choose $F_0$ in Equation (\ref{eqn:Q^0}) as a neural network which maps the feature vector $x$ to the parameters of the Gaussian $\mu(x),\sigma(x)$:
\begin{equation}
[\mu(x),\sigma(x)]^\top
= F_0 (x;\Theta_0).
\end{equation}
The conditional $\tau_k$-quantile of $\mathbf{y}$ given $\mathbf{X}=x$ is then $q^0_k(x)=\mu(x)+\sigma(x)\Phi^{-1}(\tau_k)$, where $\Phi^{-1}(\tau_k)$ is the $\tau_k$-quantile of standard Gaussian. Thus, $G_0$ in Equation (\ref{eqn:Q^0}) is
\begin{align*}
Q^0(x) & = G_0(\mu(x),\sigma(x)) \\
            & = \mu(x)+\sigma(x)[\Phi^{-1}(\tau_1),\dots,\Phi^{-1}(\tau_K)]^\top.
\end{align*}

Given the training set $\{(x_i,y_i)\}_{i\in D_{tr}}$ and the intuition of implicit regularizations, the learning objective at $t=0$ is $\min_{\Theta_0} \frac{1}{|D_{tr}|}\sum_{i\in D_{tr}} L (y_i, Q^0(x_i) )$, as in Equation (\ref{eqn:final_learning_goal}).
After this, $Q^0(x)$ may be a coarse approximation of the conditional distribution $\mathbb{P}(\mathbf{y}|\mathbf{X}=x)$. 
It is not flexible enough to express the multi-modality, asymmetry, heavy tails, etc. Carefully picking a non-Gaussian assumption relying on the expertise is far away from a good solution.
So in our approach, we improve the predictions in ensemble steps.

Notice that our framework is flexible to different application domains, because at $t=0$, it is not limited to Gaussian assumption only. One can have other choices as needed in specific tasks, for example, replacing standard Gaussian by the standard exponential distribution or some other.


\subsection{The Ensemble Steps}

Suppose $Q^{t-1}(x)$, or $q^{t-1}_{1},\dots,q^{t-1}_{K}$, obtained at step $t-1$ is known, now we give the model details in Equation (\ref{eqn:Q^t}) at the ensemble step $t\ge 1$.
It is obvious that $q^{0}_{1},\dots,q^{0}_{K}$ are strictly increasing because they obey a Gaussian distribution, so we always assume $q^{t-1}_{1},\dots,q^{t-1}_{K}$ are strictly increasing.
Specifically, when $t\ge 1$, to generate strictly increasing and better predictions $q^t_{1},\dots,q^t_{K}$, we add a new quantity to each $q^{t-1}_{k}$:
\begin{equation}
q^{t}_{k} = q^{t-1}_{k} + g^t_k(\lambda^{t}_{k}),\quad k=1,\dots,K,
\end{equation}
where $\lambda^t_k\in(-1,1)$ is a quantity that we need to determine later. 
$g^t_k(\cdot)$ is a continuous function we set that is strictly increasing over $[-1,1]$ and satisfies
\begin{align}
g^t_k(0) & = 0,\\
g^t_k(1) & = r^{t-1}_{k} := \frac{q^{t-1}_{k+1}-q^{t-1}_{k}}{2},\\
g^t_k(-1) & = l^{t-1}_{k} := \frac{q^{t-1}_{k-1}-q^{t-1}_{k}}{2}.
\end{align}
In this paper, we choose a piecewise linear $g^t_k(\cdot)$ to satisfy the above conditions:
\begin{equation}\label{eqn:g^t_k}
g^t_k(\lambda) = \left( -l^{t-1}_{k}+\mathbb{I}_{\{\lambda>0\}}\cdot(r^{t-1}_{k}+l^{t-1}_{k}) \right) \lambda.
\end{equation}

\begin{figure}[t]
\begin{center}
\begin{minipage}[t]{.36\linewidth}
\centerline{\includegraphics[width=\linewidth]{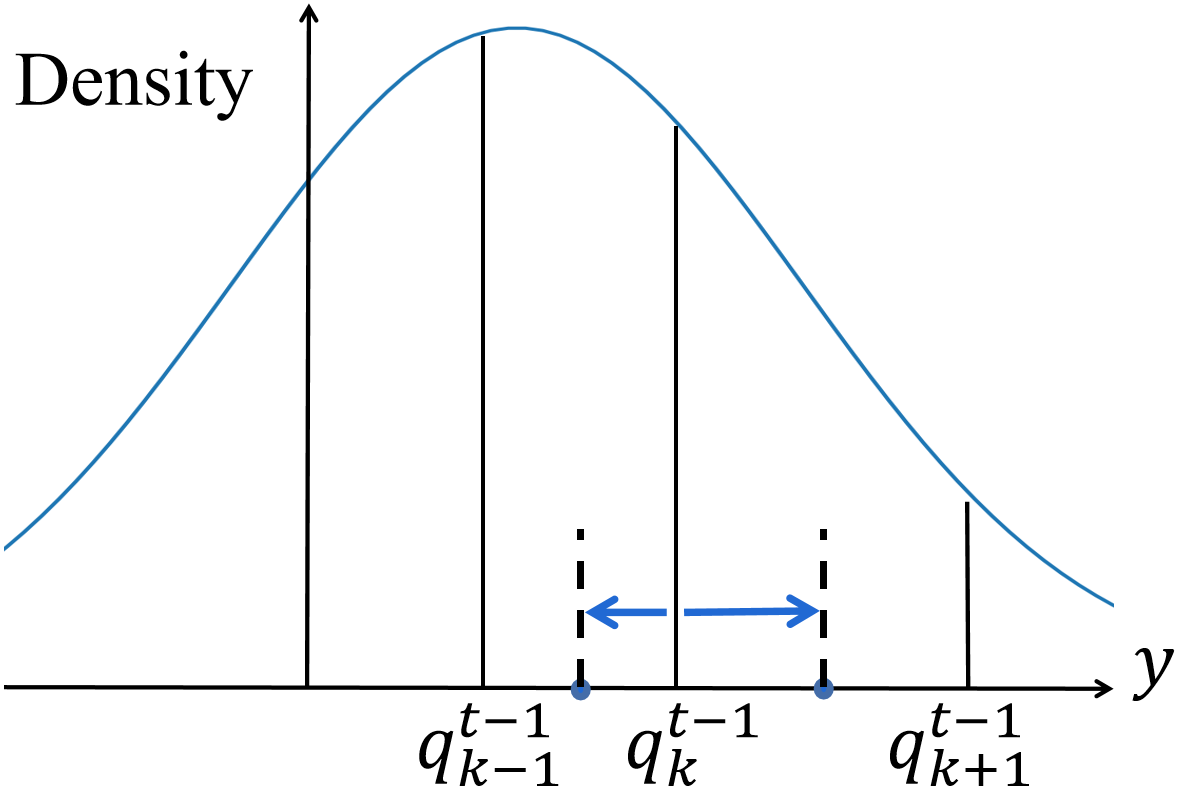}}
\end{minipage}
\hspace{2em}
\begin{minipage}[t]{.36\linewidth}
\centerline{\includegraphics[width=\linewidth]{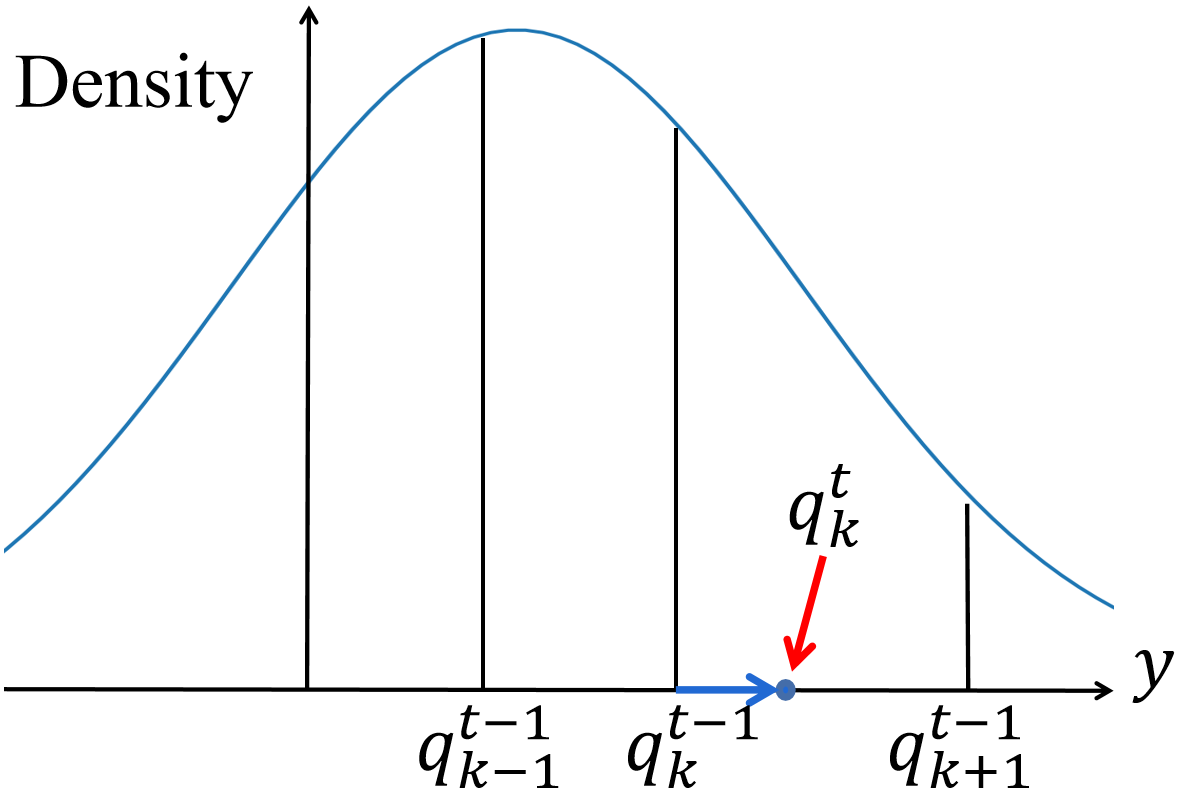}}
\end{minipage}
\caption{The illustration of the ensemble step $t\ge1$. $q^t_k$ is selected in the range between the middle point of $(q^{t-1}_{k-1},q^{t-1}_{k})$ and the middle point of $(q^{t-1}_{k},q^{t-1}_{k+1})$. By this design, $q^t_1,\dots,q^t_K$ are ensured to be strictly increasing if $q^{t-1}_1,\dots,q^{t-1}_K$ are strictly increasing.}
\label{fig:idea}
\end{center}
\vspace{-1em}
\end{figure}

To be more intuitive, one can easily verify that the left and right bounds of $q^{t}_{k}$ given $\lambda^t_k\in(-1,1)$ are the middle points of the intervals $(q^{t-1}_{k-1},q^{t-1}_{k})$ and $(q^{t-1}_{k},q^{t-1}_{k+1})$ respectively, and besides, $q^{t}_{k}=q^{t-1}_{k}$ when $\lambda^{t}_{k}=0$. This design ensures that $q^{t}_{k}$ is strictly increasing with respect to $k$ because obviously,
\begin{equation}
q^{t}_{k} < q^{t-1}_{k} + g^t_k(1) = \frac{q^{t-1}_{k+1}+q^{t-1}_{k}}{2} = q^{t-1}_{k+1} + g^{t}_{k+1}(-1) < q^{t}_{k+1}.
\end{equation}
The idea here is to move $q^{t-1}_{k}$ towards a better prediction $q^{t}_{k}$, but with only a small distance to i) remain the monotonic property, and ii) control the flexibility added in step $t$ (see the description later). We illustrate this idea in Fig. \ref{fig:idea} as well. When $k=0$ or $k=K+1$, we naturally set $q^{t-1}_0$ or $q^{t-1}_{K+1}$ to be the left/right boundary of the support set of $\mathbb{P}(\mathbf{y}|\mathbf{X}=x)$ if the boundary is finite. Now $g^t_1(\cdot)$ or $g^t_K(\cdot)$ is well-defined such that $q^{t-1}_{1}+g^t_1 (-1)$ or $q^{t-1}_{K}+g^t_{K} (1)$ is a feasible bound. Otherwise, then $q^{t-1}_0=-\infty$ or $q^{t-1}_{K+1}=+\infty$, we can now replace $q^{t-1}_0$ by $-B$ or replace $q^{t-1}_{K+1}$ by $+B$ where $B$ is a large positive constant to make $g^t_1(\cdot)$ or $g^t_K(\cdot)$ well-defined.

To make this step $t$ complete, we set $\lambda^{t}_k$ as a cubic polynomial function of $\tau_k$ followed by a tanh for transforming to $(-1,1)$, i.e., $\lambda^{t}_k = \tanh\left(\sum_{l=0}^3 a^t_l (\tau_k)^l \right)$, where $a_l^t$ is an output of the model $F_t(x;\Theta_t)$:
\begin{equation}
[a^{t}_{0},a^{t}_{1},a^{t}_{2},a^{t}_{3}]^\top
= F_t(x;\Theta_t).
\end{equation}
So far, the function $G$ in Equation (\ref{eqn:Q^t}) has been defined. We avoid letting $\lambda^{t}_k$ be totally free and be an output of $F_t(x;\Theta_t)$ directly, which will lead to totally $K$ outputs for $F_t$ (i.e., $K=99$). The cubic polynomial works as an implicit regularization that puts restrictions on $\lambda^{t}_k$, $k=1,\dots,K$. This setting controls the flexibility added in step $t$ and is crucially important and useful for adaptively balancing the distribution structure and the flexibility for $\mathbb{P}(\mathbf{y}|\mathbf{X}=x)$.

From another perspective, our approach will give a smooth density function for $\mathbb{P}(\mathbf{y}|\mathbf{X}=x)$ at each step $t$. Oppositely, letting $\lambda^{t}_k$ be totally free will give a strange or chaotic density function that cannot be interpreted, which is also the common drawback of all over-flexibility methods, as we mentioned in Section \ref{sec:motivations}. 
Back to our method, when the training dataset $\{(x_i,y_i)\}_{i\in D_{tr}}$ is given, the learning objective at step $t$ is $\min_{\Theta_t} \frac{1}{|D_{tr}|}\sum_{i\in D_{tr}} L (y_i, Q^t(x_i) )$ with implicit regularizations. We summarize the whole procedure of our approach in Algorithm \ref{alg:algorithm1}.


\begin{algorithm}[t]
\caption{\quad The ensemble multi-quantiles (EMQ) approach for uncertainty quantification.}
\label{alg:algorithm1}
\textbf{Input}: the training dataset  $\{(x_i,y_i)\}_{i\in D_{tr}}$, and $K$ probability levels of interest $0<\tau_1<\cdots<\tau_K<1$.\\
\textbf{Hyper-parameters}: the number of ensemble steps $T$; and the hyper-parameters
of the neural network models $F_0$ and $F_t$, $t\ge 1$.\\
\textbf{Output}: the learnable parameters $\Theta_0$ and $\Theta_t$, $t\ge 1$ of the models $F_0$ and $F_t$.
\begin{algorithmic}[1] 
\STATE Let $[\mu(x),\sigma(x)]^\top = F_0 (x;\Theta_0)$ and $Q^0(x) = G_0(\mu(x),\sigma(x)) = \mu(x)+\sigma(x)[\Phi^{-1}(\tau_1),\dots,\Phi^{-1}(\tau_K)]^\top$, solve:
\begin{equation}
\min_{\Theta_0} \frac{1}{|D_{tr}|}\sum_{i\in D_{tr}} L (y_i, Q^0(x_i)),
\end{equation}
where the loss function $L$ can be found in Equation (\ref{eqn:loss_L}). $\Phi^{-1}(\tau_k)$ is the $\tau_k$-quantile of standard Gaussian.
\FOR{$t = 1,\dots,T$}
\STATE Let $[a^{t}_{0},a^{t}_{1},a^{t}_{2},a^{t}_{3}]^\top
= F_t(x;\Theta_t)$, and $\lambda^{t}_{k} = \tanh\left(\sum_{l=0}^3 a^t_{l} (\tau_k)^l \right)$. Let $Q^{t}(x) = Q^{t-1}(x) + [g^t_{1}(\lambda^{t}_{1}),\dots,g^t_{K}(\lambda^{t}_{K})]^\top$, where both $g^t_{k}(\cdot)$ and $\lambda^{t}_{k}$ condition on $x$. The definition of $g^t_{k}(\cdot)$ can be found in Equation (\ref{eqn:g^t_k}). Then solve:
\begin{equation}
\min_{\Theta_t} \frac{1}{|D_{tr}|}\sum_{i\in D_{tr}} L (y_i, Q^t(x_i)).
\end{equation}
\ENDFOR
\STATE \textbf{return} the learned parameters $\Theta_t$, $t=0,1,\dots,T$.  Given a new test input $x$, the step-$T$ predictions $Q^T(x)$ are the final predictions of the quantiles of $\mathbb{P}(\mathbf{y}|\mathbf{X}=x)$.  The computing process of $Q^T(x)$ is described as above.
\end{algorithmic}
\end{algorithm}


\subsection{Adaptive $T$ Strategy}

From Algorithm \ref{alg:algorithm1}, the only hyper-parameter that is newly introduced into the EMQ approach is the number of ensemble steps $T$. Such a hyper-parameter is certainly crucial for the performance of the approach. Indeed, $T$ functions as a trade-off between the distribution structure and the flexibility, because more ensemble steps will lead to more flexible distribution that is farther away from Gaussian. Here we introduce the adaptive $T$ strategy which determines $T$ adaptively and hence can help EMQ find the balance.

The strategy works analogously to the early stopping in deep learning, thus is a kind of implicit regularization. In this strategy, we split the training dataset $D_{tr}$ into a new training dataset and a validation set (can be the same validation set used in the early stopping when training $F_t$, $t\ge 0$). The performance of the predictions $Q^t(x)$ on the validation set is tracked for all $t\ge0$ and $t\le T_{max}$. Under some criteria, the EMQ approach terminates the ensemble steps when $Q^t(x)$ starts to deteriorate. $T_{ada}$ is set to be the step $t$ at which $Q^t(x)$ is the best. The EMQ model will finally have $T_{ada}$ ensemble steps and $Q^{T_{ada}}(x)$ is the final quantile predictions. To further reduce the risk of over-fitting, in the strategy, we evaluate $Q^t(x)$ on the validation set with a metric that is different from the quantile loss $L$ in Equation (\ref{eqn:loss_L}) used for training $F_t$, $t\ge 0$.
For example, we can use the metric ECE (expected calibration error, computed by calibrating for every $\tau_k\in\Pi$ and taking the average). This adaptive $T$ strategy is summarized in Algorithm \ref{alg:algorithm2}.

\begin{algorithm}[t]
\caption{\quad The adaptive $T$ strategy.}
\label{alg:algorithm2}
\textbf{Input}: the inputs of the EMQ approach in Algorithm \ref{alg:algorithm1}; an additional validation set.\\
\textbf{Hyper-parameters}: the number of maximum ensemble steps $T_{max}$; two numbers of steps $t_1>t_2$.\\
\textbf{Output}: the adaptive number of ensemble steps $T_{ada}$.
\begin{algorithmic}[1] 
\FOR{$t = 0,\dots,T_{max}$}
\STATE Perform step $t$ of EMQ and evaluate $Q^{t}(x)$: $e_t=$ ECE of $Q^{t}(x)$ on the validation set.
\IF{$t\ge  t_1$ and avg.$(e_{t-t_2+1:t})$ $>$ avg.$(e_{t-t_1+1:t-t_2})$}
\STATE $t'=t$
\STATE Break
\ENDIF
\ENDFOR
\STATE $T_{ada}=\arg\min_{\{t:0\le t \le t'\}} e_{t}$.
\STATE \textbf{return} $T_{ada}$.  Given a new test input $x$, the final quantile predictions of EMQ are $Q^{T_{ada}}(x)$.
\end{algorithmic}
\end{algorithm}

\subsection{EMQW}

In many circumstances, people are interested in tail-side prediction intervals such as the $90\%$-interval, $80\%$-interval, etc. But the quantiles at the tail sides are more difficult to predict, so we extend EMQ to a new version which places more emphasis at the tail sides than EMQ does. The only difference between this new EMQW method and the original EMQ is that EMQW uses a weighted quantile loss function instead of Equation (\ref{eqn:loss_L}):
\begin{align}
L^{W}(y,Q^t(x)) = \sum_{k=1}^K w_{\tau_k} L_{\tau_k} (y,q^t_k(x)), \label{eqn:loss_EMQW}\\
L_{\tau} (y,q) =  (y-q)(\tau-\mathbb{I}_{\{y<q\}}), \label{eqn:loss_EMQW_tau}\\
w_\tau = 1/\mathbb{E}_{y\sim\mathcal{N}(0,1)}\left[L_{\tau} (y,\Phi^{-1}(\tau))\right].  \label{eqn:weights_EMQW_tau}
\end{align}
$\Phi$ is the  cumulative distribution function of $\mathcal{N}(0,1)$. This aims to let the losses for different $\tau_k$ have a similar magnitude, because $\mathbb{E}_{y\sim\mathcal{N}(0,1)}\left[L_{\tau} (y,\Phi^{-1}(\tau))\right]$ may be quite different for different $\tau$. This was also suggested by \cite{bondell2010noncrossing,liu2011simultaneous}.
Except for this difference, EMQW and EMQ share the same designs and settings. We will show in the experiments that EMQW is a better choice if tail-side quantiles should be considered, and we will use it to compare to other methods.

\subsection{Computational Complexity}

At step 0, the computational complexity of training $F_0$ is the same as that of training an ordinary multi-layer perception with a 2-dimensional output layer. We denote the computational cost of training $F_0$ as $C_0$.
At step $t>0$, the network model $F_t$ has a much smaller structure than $F_0$ does. So, training $F_t$ takes much less time. If the computational cost of training $F_t$ is $C_1$ (then $C_1\ll C_0$), the total computational cost of training EMQ or EMQW will be $C_0+T_{ada}C_1$, and at most $C_0+T_{max}C_1$. We will see in the experimental section that setting $T_{max}=40$ is enough for most datasets. 

For space complexity, as we should store the predicted quantiles at every step, the additional space we need except for that consumed in training neural networks is $O(|D_{tr}|\cdot |\Pi| \cdot T_{ada})$, and at most $O(|D_{tr}|\cdot |\Pi| \cdot T_{max})$ ($|\Pi|=99$ in our experiments). It is noticeable that training $T_{ada}$ networks requires only the same space as training $F_0$ does because $F_0$ is the biggest one.

\section{Experiments}

In this section, we implement the EMQ approach and its variant EMQW, and compare it with both recent popular and classical methods on a wide range of datasets. The extensive evaluations reveal that EMQW achieves state-of-the-art performance. We also visualize the learning process and learning results of EMQW for better interpretation, and demonstrate that empirically it works as we expect. We show that it captures a variety of distribution shapes existing in real data. Consequently, we can conclude that the proposed approach owns the merits we state before.

\subsection{Datasets and Evaluation Settings}

We collect 20 UCI datasets \cite{Dua:2019} and show their information in Table \ref{tab:data-info}. The datasets are from different domains and can be used for comprehensive regression studies. For every dataset, we extract a short name from its full name. The numbers of samples vary from 10,000 to 515,345 and the feature dimensions vary from 5 to 465. The 20 datasets cover both medium-size and large-size datasets, with both low-dimensional and high-dimensional features.

For evaluation, we randomly select 20\% from the full dataset as the testing set, and do the training and testing $N_{te}$ times with different training/testing splits, for averaging the evaluation metrics. We set $N_{te}=5$ for the datasets with sample size less then 100,000 and $N_{te}=1$ for the rest: GPU, Query, Wave, Air, and Year for reducing the computational burden. In addition, the proportion of the testing set is raised from 20\% to 50\% for the two largest datasets Air and Year, for further reducing the computational burden. All features and labels are normalized to have sample mean 0 and sample variance 1.

\begin{table}[t]
  \centering
  \caption{The information of the 20 UCI datasets used.}
  \begin{tabular}{|c|c|c|c|}
  \Xhline{1pt}
\makecell{Dataset\\Name} & Full Name & \makecell{Sample\\Size} & \makecell{Feature\\Dimension} \\ \Xhline{1pt}
Grid & Electrical Grid Stability & 10,000 & 12\\ \hline
Naval & \makecell{Conditional Based\\Maintenance \cite{coraddu2016machine}} & 11,933 & 15\\ \hline
Appliance & \makecell{Appliances energy\\prediction \cite{candanedo2017data}} & 19,735 & 27\\ \hline
Election & Real-time Election \cite{moniz2019real} & 21,643 & 23\\ \hline
Steel & \makecell{Steel Industry Energy\\Consumption} & 35,040 & 6\\ \hline
Facebook1 & \makecell{Facebook Comment\\Volume 1 \cite{singh2015comment}} & 40,949 & 52\\ \hline
Facebook2 & \makecell{Facebook Comment\\Volume 2 \cite{singh2015facebook}} & 81,312 & 52\\ \hline
PM2.5 & Beijing PM2.5 \cite{liang2015assessing} & 41,757 & 10\\ \hline
Bio & \makecell{Physicochemical\\Properties} & 45,730 & 9\\ \hline
Blog & BlogFeedback \cite{buza2013feedback} & 52,397 & 276\\ \hline
Consumption & \makecell{Power consumption\\of Tetouan city \cite{salam2018comparison}} & 52,416 & 5\\ \hline
Video & \makecell{Online Video\\Characteristics} & 68,784 & 16\\ \hline
GPU & \makecell{GPU kernel\\performance \cite{ballester2019sobol}} & 241,600 & 14\\ \hline
Query & \makecell{Query Analytics\\Workloads \cite{savva2018explaining,anagnostopoulos2018scalable}}  & 199,843 & 6\\ \hline
Wave & \makecell{Wave Energy\\Converters} & 287,999 & 48\\ \hline
Air & \makecell{Beijing Multi-Site\\Air-Quality Data \cite{zhang2017cautionary}} & 383,585 & 14\\ \hline
Year & YearPredictionMSD & 515,345 & 90\\ \hline
Relative & \makecell{Relative location\\of CT slices} & 53,500 & 379\\ \hline
Ujiindoorloc & UJIIndoorLoc \cite{torres2014ujiindoorloc} & 19,937 & 465\\ \hline
Cup98 & KDD Cup 1998 & 95,412 & 326\\ \Xhline{1pt}
  \end{tabular}
  \label{tab:data-info}
\end{table}

\subsection{Evaluation Metrics}

To evaluate the predicted quantiles of $\mathbb{P}(\mathbf{y}|\mathbf{X}=x)$, we use both calibration and sharpness. The probability level set is $\Pi=\{\tau_k\}_{k=1}^K=\{0.01,0.02,\dots,0.98,0.99\}$ in this paper, with totally $K=99$ levels. Given the testing set $\{(x_i,y_i)\}_{i\in D_{te}}$ and a machine learning model that is ready to predict, we temporarily denote the predicted $\tau$-quantile of $\mathbb{P}(\mathbf{y}|\mathbf{X}=x_i)$ as $q_\tau (x_i)$, for any $i\in D_{te}$ and $\tau\in\Pi$. Three evaluation metrics (times 100 in our tables) are used:

\textbf{i) EICE (expected interval calibration error)}. Generally in uncertainty quantification, the construction of intervals is required. We compute the calibration error of the interval constructed by each pair of quantiles, e.g., $[q_{0.1} (x_i),q_{0.9} (x_i)]$, and take the average over all intervals. To be short,
\begin{equation}
    \text{EICE} = \frac{1}{49}\sum_{k=1}^{49}  \bigg|  1-2\tau_k  -  \frac{1}{|D_{te}|}\sum_{i\in{D_{te}}}  \mathbb{I}_{\{q_{\tau_k} (x_i) < y_i < q_{\tau_{100-k}} (x_i)\}}  \bigg|.
\end{equation}

\textbf{ii) EIS (expected interval sharpness)}. Sharpness measures the length of the intervals and is an evaluation metric supplement to calibration, because good calibration can be obtained by unsatisfactory intervals such as those from the unconditional distribution of $\mathbf{y}$. We define 
\begin{equation}
    \text{EIS} = \frac{1}{49}\sum_{k=1}^{49} \frac{1}{|D_{te}|} \sum_{i\in{D_{te}}}  \big| q_{\tau_{100-k}} (x_i) - q_{\tau_k} (x_i) \big|.
\end{equation}

\textbf{iii) TICE (tail interval calibration error)}. Researchers and practitioners may be more interested in the 90\%-interval, 80\%-interval, etc. So we use only the intervals constructed by quantiles at the tail sides, and define 
\begin{equation}
    \text{TICE} = \frac{1}{|\Pi_{ta}|}\sum_{\tau\in\Pi_{ta}}  \bigg|  1-2\tau  -  \frac{1}{|D_{te}|}\sum_{i\in{D_{te}}}  \mathbb{I}_{\{q_{\tau} (x_i) < y_i < q_{1-\tau} (x_i)\}}  \bigg|,
\end{equation}
where $\Pi_{ta}=\{0.05,0.1,0.15,0.2\}$. It corresponds to the average calibration of  90\%,  80\%,  70\%,  and 60\%-intervals.


\subsection{Methods and Implementations}

We implement EMQ and two variants, as well as 8 popular existing deep learning-based methods for uncertainty quantification for comparisons. 
In them, HNN and Deep Ensemble are variance network models. MC Dropout and Concrete Dropout are Bayesian methods. Vanilla QR, QRW, and SQR are quantile regression models. Interval Score is a recent method related to quantile regression.
The implementation details of every method are summarized as follows.

\textbf{EMQ}. In EMQ implementation, we set the only additional hyper-parameter $T_{max}=40$ if the dataset has a feature dimension less than 300, and $T_{max}=200$ otherwise. For the network structure, at the step $t=0$ for $F_0$, we use 3 hidden layers with layer sizes $[8,16,4]\times d$ where $d$ is the input dimension. The output dimension is 2 at $t=0$.
At the step $t>0$ for $F_t$, we use 2 hidden layers with layer sizes $[16,8]$, which are much smaller than those in $F_0$. Now the output dimension is 4 at $t>0$. For activation functions, we use tanh for all hidden layers and linear for all output layers. Softplus is applied additionally for ensuring a positive output if necessary.

\textbf{EMQ$_\mathbf{0}$}. We also implement a special case of EMQ in which we set $T_{max}=0$, named EMQ$_0$. It is equivalent to assuming $\mathbb{P}(\mathbf{y}|\mathbf{X}=x)$ to be Gaussian all along, without further ensemble steps. We can use it to prove the advantages of EMQ.
The hyper-parameters of EMQ$_0$ are set the same as EMQ (at the step $t=0$).

\textbf{EMQW}. Except for the loss function, EMQW and EMQ share the same hyper-parameters. We aim to learn better intervals at the tail sides, and compare with other competing methods using this EMQW.

\textbf{HNN}. We train a mean network and a variance network alternately twice with negative log-likelihood loss, as suggested by \cite{skafte2019reliable}. Two hidden layers with layer sizes $[8,4]\times d$ are used for reducing the computational cost, because HNN is the building block of Deep Ensemble whose computational complexity is the highest among all methods as it requires training multiple HNNs. Only alternately training twice is also due to this reason. For the mean network, the ReLU activation is used on hidden layers and linear on output layer, and for the variance network, tanh is used on hidden layers and Softplus on output layer.

\textbf{Deep Ensemble}. We train 5 HNNs with different initialization seeds for ensemble. The hyper-parameters of every HNN are exactly the same as described above. A mixture of Gaussian combining the obtained multiple Gaussian distributions is assumed and its mean and variance are computed to form an approximated Gaussian, as done in \cite{lakshminarayanan2017simple}.

\textbf{MC Dropout}. The network in MC Dropout has 3 hidden layers with sizes $[8,16,4]\times d$ and the dropout probability is $0.1$, as suggested in \cite{gal2016dropout}. ReLU is applied on hidden layers and linear on output layer. Mean squared error is used for training the network. When predicting, 1,000 predictions for every input under the training mode are obtained for computing the mean and variance.

\textbf{Concrete Dropout}. The network in Concrete Dropout has 2 hidden layers with sizes $[8,4]\times d$. We use the official code released in \cite{gal2017concrete}. If in some cases the network cannot give a valid training loss, we change the hidden layer sizes to be the suggested $[100,80]$ in the referenced code.

\textbf{Vanilla QR}. In vanilla quantile regression, we use a network having 3 hidden layers with sizes $[8,16,4]\times d$, which outputs the 99 conditional quantiles directly. ReLU is applied on hidden layers and linear on output layer. The quantile loss in Equation (\ref{eqn:loss_L}) is used to train the network.

\textbf{QRW}. In this new method called quantile regression with weights (QRW), we use the same network as in Vanilla QR. The only difference is that the weighted quantile loss in Equation (\ref{eqn:loss_EMQW}) is used instead of the un-weighted version in Equation (\ref{eqn:loss_L}). Similarly, this will give more emphasis on the tail sides than Vanilla QR does.

\textbf{SQR}. In simultaneous quantile regression \cite{tagasovska2019single}, we use the hidden layer sizes $[8,16,4]\times d$ for the network. ReLU is applied on hidden layers and linear on output layer. The quantile loss of a random $\tau$ of every data point is used to update the network parameters at each descent step. After training, the quantiles of any levels can be predicted.

\textbf{Interval Score}. In Interval Score method, we use the hidden layer sizes $[8,16,4]\times d$ for the network too. It outputs the 99 conditional quantiles directly as well. The only difference to Vanilla QR is that now the interval score in Equation (\ref{eqn:interval_score}) is adopted to train the network.

\textbf{Common settings}. For all the above methods, we use some common settings for all neural network training. The batch size is 128 and the learning rate is 0.01. We randomly select 20\% from the training set as the validation set, which is used in the early stopping strategy. The maximum number of epochs is 1,000. All the optimization is done with Adam \cite{kingma2014adam} implemented by PyTorch \cite{NEURIPS2019_9015}. Initializations of all neural networks are done by Xavier Normal method \cite{glorot2010understanding} with specific seeds for reproducibility.

\begin{table*}[t]
\begin{minipage}{0.32\linewidth}
  \centering
  \caption{EICE results of the three models in EMQ family. EMQW wins on 11 datasets, EMQ wins on 5, and EMQ$_0$ wins on 4. $^*$ is the best among three.}
  \begin{tabular}{|c|c|c|c|}
  \Xhline{1pt}
Dataset & EMQ$_0$ & EMQ & EMQW \\ \Xhline{1pt}
Grid & 5.33 & 5.93 & $^*$3.33 \\ \hline
Naval & 18.95 & $^*$2.47 & 2.87 \\ \hline
Appliance & 2.27 & 2.26 & $^*$1.99 \\ \hline
Election & 27.22 & 24.62 & $^*$22.21 \\ \hline
Steel & 18.77 & 9.92 & $^*$8.81 \\ \hline
Facebook1 & 8.18 & 4.88 & $^*$3.03 \\ \hline
Facebook2 & 7.57 & $^*$3.26 & 3.74 \\ \hline
PM2.5 & 1.69 & 1.90 & $^*$1.05 \\ \hline
Bio & 1.85 & 1.98 & $^*$1.10 \\ \hline
Blog & 11.57 & $^*$2.90 & 3.34 \\ \hline
Consumption & $^*$1.46 & 1.52 & 2.05 \\ \hline
Video & 9.93 & 6.33 & $^*$3.85 \\ \hline
GPU & 3.61 & 3.54 & $^*$1.67 \\ \hline
Query & $^*$2.15 & 2.15 & 3.50 \\ \hline
Wave & $^*$3.34 & 3.34 & 3.80 \\ \hline
Air & 1.37 & $^*$0.84 & 1.21 \\ \hline
Year & $^*$0.54 & 1.12 & 1.77 \\ \hline
Relative & 6.78 & 6.48 & $^*$3.86 \\ \hline
Ujiindoorloc & 7.42 & 4.37 & $^*$3.11 \\ \hline
Cup98 & 27.07 & $^*$4.46 & 13.87 \\ \Xhline{1pt}
  \end{tabular}
  \label{tab:interval_ECE_EMQ_family}
  \end{minipage}\hfill
  \begin{minipage}{0.33\linewidth}
  \centering
  \caption{EIS results of the three models in EMQ family. EMQW wins on 3 datasets, EMQ wins on 13, and EMQ$_0$ wins on 4.}
  \begin{tabular}{|c|c|c|c|}
  \Xhline{1pt}
Dataset & EMQ$_0$ & EMQ & EMQW \\ \Xhline{1pt}
Grid & 21.52 & $^*$20.26 & 20.94 \\ \hline
Naval & 19.39 & $^*$9.13 & 10.26 \\ \hline
Appliance & 80.41 & $^*$78.23 & 88.67 \\ \hline
Election & 8.42 & $^*$6.34 & 8.66 \\ \hline
Steel & 2.24 & $^*$1.45 & 1.51 \\ \hline
Facebook1 & 21.63 & $^*$20.70 & 24.73 \\ \hline
Facebook2 & 22.68 & $^*$19.50 & 23.26 \\ \hline
PM2.5 & 51.77 & $^*$50.74 & 54.35 \\ \hline
Bio & $^*$69.91 & 70.08 & 80.51 \\ \hline
Blog & 45.04 & $^*$36.49 & 42.60 \\ \hline
Consumption & 92.63 & $^*$90.62 & 90.98 \\ \hline
Video & $^*$32.43 & 51.92 & 60.72 \\ \hline
GPU & 6.20 & 5.82 & $^*$5.81 \\ \hline
Query & $^*$9.35 & 9.35 & 9.44 \\ \hline
Wave & $^*$10.00 & 10.00 & 12.09 \\ \hline
Air & 28.25 & $^*$28.06 & 29.90 \\ \hline
Year & 107.60 & $^*$103.81 & 106.32 \\ \hline
Relative & 39.25 & 37.55 & $^*$20.56 \\ \hline
Ujiindoorloc & 53.30 & 33.11 & $^*$22.47 \\ \hline
Cup98 & 39.47 & $^*$27.25 & 39.57 \\ \Xhline{1pt}
  \end{tabular}
  \label{tab:interval_sharpness_EMQ_family}
  \end{minipage}\hfill
  \begin{minipage}{0.32\linewidth}
  \centering
  \caption{TICE results of the three models in EMQ family. EMQW wins on 12 datasets, EMQ wins on 5, and EMQ$_0$ wins on 3.}
  \begin{tabular}{|c|c|c|c|}
  \Xhline{1pt}
Dataset & EMQ$_0$ & EMQ & EMQW \\ \Xhline{1pt}
Grid & 7.21 & 8.49 & $^*$4.72 \\ \hline
Naval & 18.94 & $^*$2.30 & 2.66 \\ \hline
Appliance & 2.97 & 3.59 & $^*$2.00 \\ \hline
Election & 18.60 & 18.63 & $^*$14.69 \\ \hline
Steel & 18.93 & 11.42 & $^*$10.59 \\ \hline
Facebook1 & 7.61 & 4.74 & $^*$3.43 \\ \hline
Facebook2 & 7.42 & $^*$3.43 & 3.62 \\ \hline
PM2.5 & 2.61 & 3.14 & $^*$1.74 \\ \hline
Bio & 2.57 & 3.29 & $^*$1.30 \\ \hline
Blog & 9.32 & $^*$3.23 & 4.43 \\ \hline
Consumption & $^*$1.87 & 2.18 & 2.86 \\ \hline
Video & 8.31 & 6.34 & $^*$4.70 \\ \hline
GPU & 3.45 & 3.29 & $^*$1.34 \\ \hline
Query & 3.40 & 3.40 & $^*$2.41 \\ \hline
Wave & $^*$3.29 & 3.29 & 4.61 \\ \hline
Air & 2.14 & $^*$1.32 & 1.96 \\ \hline
Year & $^*$0.68 & 1.72 & 1.95 \\ \hline
Relative & 6.74 & 5.73 & $^*$4.81 \\ \hline
Ujiindoorloc & 6.68 & 4.23 & $^*$3.91 \\ \hline
Cup98 & 18.72 & $^*$5.17 & 10.16 \\ \Xhline{1pt}
  \end{tabular}
  \label{tab:tail_interval_ECE_EMQ_family}
  \end{minipage}
\end{table*}

\subsection{Performance of EMQ Family}

We first examine the three models in EMQ family: EMQ$_0$, EMQ, and EMQW. Intuitively, EMQ$_0$ assumes conditional Gaussian and lacks flexibility, EMQ adds ensemble steps for seeking a balance between the distribution structure and the flexibility, and EMQW places more weights on the tail-side quantiles. We report their performance under the three evaluation metrics respectively in Table \ref{tab:interval_ECE_EMQ_family}, \ref{tab:interval_sharpness_EMQ_family}, and \ref{tab:tail_interval_ECE_EMQ_family}.

From the results reported, we find that EMQ significantly outperforms EMQ$_0$ on both calibration and sharpness (EMQ beats EMQ$_0$ on 15 datasets in Table \ref{tab:interval_ECE_EMQ_family}, and beats EMQ$_0$ on 18 in Table \ref{tab:interval_sharpness_EMQ_family}), verifying that the ensemble steps in EMQ successfully add appropriate flexibility into the conditional distributions learned. 
EMQW performs further better than EMQ on calibration, or on both EICE and TICE metrics. It indeed gives better tail-side intervals, as we expect.
On the sharpness metric EIS, EMQW performs slightly worse than EMQ, but relatively the results are close. Overall, we think EMQW is the best choice in the three, and in the following experiments, we will use EMQW as our model choice when comparing to other methods.

\subsection{Performance of Competing Methods}

\begin{table*}[t]
  \centering
  \caption{EICE results of nine methods. $^{**}$ is the best performance and $^*$ is the second best. EMQW wins the best on 13 out of 20 datasets, and wins the second best or above on 16 out of 20 datasets. On some datasets, the improvements over other methods are large (decrease rate over 50\%), such as Facebook1, Facebook2, Bio, Blog, Video, GPU, Wave, and Cup98.}
  \begin{tabular}{|c|c|c|c|c|c|c|c|c|c|}
  \Xhline{1pt}
Dataset $\backslash$ Method & EMQW & HNN & \makecell{Deep\\Ensemble} & MC Dropout & \makecell{Concrete\\Dropout} & Vanilla QR & QRW & SQR & \makecell{Interval\\Score} \\ \Xhline{1pt}
Grid & 3.33 & 7.68 & 13.07 & $^{**}$2.03 & $^*$2.41 & 6.14 & 4.54 & 4.55 & 5.32 \\ \hline
Naval & $^{**}$2.87 & $^*$2.97 & 15.62 & 17.48 & 11.02 & 14.46 & 11.97 & 15.23 & 12.36 \\ \hline
Appliance & $^{**}$1.99 & 3.37 & 8.40 & 21.99 & 6.05 & 4.58 & 2.88 & 5.48 & $^*$2.38 \\ \hline
Election & 22.21 & $^{**}$8.62 & 23.62 & 21.90 & $^*$21.34 & 32.63 & 29.87 & 34.65 & 30.49 \\ \hline
Steel & $^*$8.81 & $^{**}$8.28 & 22.11 & 28.88 & 22.68 & 16.79 & 14.58 & 18.65 & 16.75 \\ \hline
Facebook1 & $^{**}$3.03 & $^*$15.99 & 23.17 & 24.97 & 26.93 & 17.06 & 16.74 & 25.76 & 18.44 \\ \hline
Facebook2 & $^{**}$3.74 & 18.90 & 25.33 & 17.15 & 34.26 & 17.50 & 17.54 & 25.96 & $^*$16.48 \\ \hline
PM2.5 & $^{**}$1.05 & $^*$1.71 & 8.24 & 16.28 & 2.04 & 3.32 & 3.37 & 5.92 & 2.82 \\ \hline
Bio & $^{**}$1.10 & 3.30 & 7.77 & 24.35 & 2.05 & 2.55 & 2.76 & $^*$1.97 & 2.41 \\ \hline
Blog & $^{**}$3.34 & 28.07 & 37.39 & 31.42 & 27.33 & 22.23 & 22.57 & 49.47 & $^*$21.50 \\ \hline
Consumption & 2.05 & 1.93 & $^{**}$1.14 & 32.41 & 1.95 & 2.04 & $^*$1.15 & 2.11 & 1.51 \\ \hline
Video & $^{**}$3.85 & 21.47 & 24.68 & 10.45 & 18.92 & 8.39 & 7.59 & 18.11 & $^*$7.14 \\ \hline
GPU & $^{**}$1.67 & $^*$3.88 & 11.99 & 10.76 & 8.70 & 7.62 & 5.45 & 6.85 & 6.52 \\ \hline
Query & $^*$3.50 & $^{**}$1.02 & 13.10 & 14.96 & 4.44 & 3.70 & 4.38 & 3.63 & 5.56 \\ \hline
Wave & $^{**}$3.80 & 28.25 & 14.81 & 39.23 & 21.48 & 20.76 & $^*$11.81 & 27.73 & 17.88 \\ \hline
Air & $^{**}$1.21 & 3.53 & 7.72 & 7.18 & 2.70 & 2.13 & 1.92 & $^*$1.83 & 2.63 \\ \hline
Year & $^*$1.77 & 2.88 & 5.02 & 37.34 & 24.99 & $^{**}$1.57 & 2.11 & 3.38 & 1.84 \\ \hline
Relative & 3.86 & $^{**}$1.85 & 20.64 & $^*$2.46 & 16.85 & 4.98 & 3.30 & 8.20 & 3.77 \\ \hline
Ujiindoorloc & $^{**}$3.11 & 9.37 & 27.30 & 6.54 & 21.47 & 4.49 & 3.36 & 11.68 & $^*$3.25 \\ \hline
Cup98 & $^{**}$13.87 & $^*$30.36 & 34.45 & 48.43 & 31.14 & 33.78 & 34.05 & 50.00 & 33.83 \\ \Xhline{1pt}
  \end{tabular}
  \label{tab:interval_ECE}
\end{table*}

\begin{table*}[t]
  \centering
  \caption{EIS results of nine methods. $^{**}$ is the best performance and $^*$ is the second best. Some methods give unusual sharpness results, e.g., MC Dropout  or SQR gives much smaller sharpness and Concrete Dropout gives infinity on several datasets. Excluding these, the remaining methods give comparable sharpness results.}
  \begin{tabular}{|c|c|c|c|c|c|c|c|c|c|}
  \Xhline{1pt}
Dataset $\backslash$ Method & EMQW & HNN & \makecell{Deep\\Ensemble} & MC Dropout & \makecell{Concrete\\Dropout} & Vanilla QR & QRW & SQR & \makecell{Interval\\Score} \\ \Xhline{1pt}
Grid & 20.94 & $^{**}$16.68 & 21.30 & 23.97 & 18.83 & $^*$17.38 & 18.52 & 17.71 & 19.72 \\ \hline
Naval & 10.26 & $^{**}$6.31 & $^*$8.85 & 27.37 & 1349.80 & 13.60 & 12.72 & 14.83 & 12.31 \\ \hline
Appliance & 88.67 & 84.52 & 93.84 & $^{**}$35.33 & 105.06 & 69.77 & 75.12 & $^*$67.77 & 77.79 \\ \hline
Election & 8.66 & 5.10 & 6.75 & 11.51 & 54749.11 & 2.41 & $^*$2.39 & $^{**}$1.64 & 2.49 \\ \hline
Steel & $^{**}$1.51 & 2.30 & 3.16 & 16.10 & 4.73 & 2.28 & 1.86 & 1.95 & $^*$1.81 \\ \hline
Facebook1 & 24.73 & 38.24 & 40.72 & $^{**}$15.77 & inf & 18.88 & 20.50 & $^*$16.57 & 23.54 \\ \hline
Facebook2 & 23.26 & 40.68 & 45.68 & $^*$19.13 & inf & 19.61 & 20.43 & $^{**}$16.62 & 20.27 \\ \hline
PM2.5 & 54.35 & 65.67 & 71.83 & $^{**}$35.90 & 81.27 & 49.74 & 53.22 & $^*$49.33 & 53.27 \\ \hline
Bio & 80.51 & 91.93 & 97.79 & $^{**}$34.55 & 752080.32 & $^*$73.23 & 73.62 & 77.81 & 76.35 \\ \hline
Blog & 42.60 & 87.43 & 107.62 & 25.14 & inf & $^*$23.67 & 27.44 & $^{**}$0.71 & 28.49 \\ \hline
Consumption & $^*$90.98 & 95.77 & 98.42 & $^{**}$25.52 & 100.24 & 91.57 & 94.88 & 98.77 & 95.29 \\ \hline
Video & 60.72 & 86.51 & 89.82 & $^{**}$19.07 & inf & 53.30 & 52.54 & $^*$50.13 & 54.34 \\ \hline
GPU & 5.81 & 5.26 & 6.09 & 15.56 & 26.18 & $^*$4.09 & $^{**}$3.53 & 5.14 & 4.68 \\ \hline
Query & 9.44 & 10.29 & 11.96 & 21.85 & 17.37 & $^{**}$8.16 & 9.30 & 9.41 & $^*$8.64 \\ \hline
Wave & 12.09 & $^{**}$0.02 & 5.46 & 16.46 & 8.38 & $^*$0.43 & 0.50 & 0.47 & 0.49 \\ \hline
Air & 29.90 & 29.76 & 30.76 & $^{**}$21.71 & 35.59 & $^*$23.75 & 24.53 & 25.77 & 24.62 \\ \hline
Year & 106.32 & 112.28 & 116.45 & $^{**}$21.03 & inf & 105.59 & 105.05 & $^*$93.42 & 101.93 \\ \hline
Relative & 20.56 & $^*$10.08 & 37.44 & 18.23 & 179089.01 & 43.86 & 44.50 & $^{**}$7.16 & 21.75 \\ \hline
Ujiindoorloc & 22.47 & 69.39 & 122.46 & 37.06 & inf & $^*$17.28 & 24.24 & $^{**}$10.06 & 31.29 \\ \hline
Cup98 & 39.57 & 141.73 & 149.39 & $^*$6.16 & 528.60 & 26.74 & 26.17 & $^{**}$0.00 & 26.30 \\ \Xhline{1pt}
  \end{tabular}
  \label{tab:interval_sharpness}
\end{table*}

\begin{table*}[t]
  \centering
  \caption{TICE results of nine methods. They correspond to the average calibration of  90\%,  80\%,  70\%,  and 60\%-intervals. $^{**}$ is the best performance and $^*$ is the second best. EMQW wins the best on 10 out of 20 datasets, and wins the second best or above on 16 out of 20 datasets. On some datasets, the improvements over other methods are large (decrease rate close to or over 50\%), such as Facebook1, Facebook2, Blog, GPU, Wave, and Cup98.}
  \begin{tabular}{|c|c|c|c|c|c|c|c|c|c|}
  \Xhline{1pt}
Dataset $\backslash$ Method & EMQW & HNN & \makecell{Deep\\Ensemble} & MC Dropout & \makecell{Concrete\\Dropout} & Vanilla QR & QRW & SQR & \makecell{Interval\\Score} \\ \Xhline{1pt}
Grid & 4.72 & 10.88 & 13.14 & $^{**}$1.74 & $^*$2.99 & 7.02 & 5.51 & 5.68 & 5.98 \\ \hline
Naval & $^*$2.66 & $^{**}$2.46 & 17.89 & 18.53 & 12.56 & 15.11 & 16.23 & 11.93 & 14.79 \\ \hline
Appliance & $^{**}$2.00 & $^*$2.66 & 8.02 & 32.18 & 6.18 & 5.82 & 3.29 & 8.97 & 3.30 \\ \hline
Election & $^*$14.69 & $^{**}$6.66 & 23.08 & 18.49 & 19.18 & 32.72 & 25.69 & 52.57 & 26.10 \\ \hline
Steel & $^*$10.59 & $^{**}$6.06 & 23.07 & 20.47 & 19.73 & 13.93 & 13.75 & 15.54 & 11.76 \\ \hline
Facebook1 & $^{**}$3.43 & 13.34 & 18.64 & 36.87 & 18.18 & 17.89 & $^*$8.71 & 39.24 & 15.15 \\ \hline
Facebook2 & $^{**}$3.62 & 14.69 & 19.21 & 23.06 & 22.83 & 10.83 & $^*$10.22 & 39.28 & 10.25 \\ \hline
PM2.5 & $^{**}$1.74 & $^*$1.80 & 10.57 & 24.02 & 3.13 & 4.54 & 3.17 & 6.64 & 3.21 \\ \hline
Bio & $^{**}$1.30 & 2.61 & 7.33 & 35.70 & 3.55 & 3.28 & 3.36 & $^*$1.72 & 2.34 \\ \hline
Blog & $^{**}$4.43 & 17.93 & 22.13 & 46.60 & 21.53 & 18.78 & 22.22 & 74.19 & $^*$10.10 \\ \hline
Consumption & 2.86 & 2.19 & 1.97 & 48.20 & 2.21 & 1.97 & $^{**}$1.15 & 2.94 & $^*$1.41 \\ \hline
Video & $^{**}$4.70 & 14.15 & 15.60 & 14.36 & 13.77 & 6.12 & $^*$5.77 & 26.52 & 6.69 \\ \hline
GPU & $^{**}$1.34 & $^*$2.71 & 16.04 & 13.05 & 9.26 & 5.75 & 4.51 & 5.09 & 4.70 \\ \hline
Query & $^*$2.41 & $^{**}$0.76 & 15.57 & 15.39 & 5.31 & 4.93 & 3.43 & 4.86 & 8.55 \\ \hline
Wave & $^{**}$4.61 & 20.69 & 16.25 & 25.00 & 21.67 & 15.98 & $^*$7.59 & 23.86 & 14.83 \\ \hline
Air & 1.96 & 3.80 & 8.88 & 12.81 & 3.62 & $^*$1.11 & 1.35 & $^{**}$1.08 & 3.08 \\ \hline
Year & $^*$1.95 & 4.04 & 6.78 & 55.60 & 19.27 & $^{**}$1.43 & 2.59 & 4.58 & 1.95 \\ \hline
Relative & 4.81 & $^{**}$1.59 & 20.66 & $^*$2.31 & 14.25 & 5.06 & 3.83 & 6.69 & 3.29 \\ \hline
Ujiindoorloc & $^*$3.91 & 5.93 & 22.75 & 7.06 & 17.73 & 6.50 & 4.10 & 19.69 & $^{**}$3.12 \\ \hline
Cup98 & $^{**}$10.16 & $^*$18.73 & 20.42 & 72.65 & 20.48 & 26.93 & 26.67 & 75.00 & 25.69 \\ \Xhline{1pt}
  \end{tabular}
  \label{tab:tail_interval_ECE}
\end{table*}

We compare the proposed EMQW to other competing methods in Table \ref{tab:interval_ECE}, \ref{tab:interval_sharpness}, and \ref{tab:tail_interval_ECE} which show EICE, EIS, and TICE results respectively. 
The competing methods include variance networks, Bayesian methods, quantile regression-based models, etc., which are extensive and highly representative.

From EICE results in Table \ref{tab:interval_ECE}, we can find that EMQW wins the best on 13 out of 20 datasets, and wins the second best or above on 16 out of 20 datasets. On some datasets where EMQW wins the best, such as Facebook1, Facebook2, Bio, Blog, Video, GPU, Wave, and Cup98, the improvements over other methods are large, producing a decrease rate of about 50\% or even 80\%. Conclusively, EMQW produces superior state-of-the-art performance on the predicted quantiles spreading the whole distribution. 

From EIS results in Table \ref{tab:interval_sharpness}, we can find that some methods give unusual sharpness results. MC Dropout and SQR give very small sharpness. However,  their calibration results are unacceptable.
Concrete Dropout has no good performance on sharpness, and on calibration as well. Excluding these, the remaining methods have comparable performance on the sharpness. 
Thus, for the remaining methods, the two calibration metrics EICE and TICE are more important now. Our EMQW achieves an appropriate balance between the calibration and sharpness.

From TICE results in Table \ref{tab:tail_interval_ECE}, we can find that EMQW wins the best on 10 out of 20 datasets, and wins the second best or above on 16 out of 20 datasets. On some datasets where EMQW wins the best, such as Facebook1, Facebook2, Blog, GPU, Wave, and Cup98, the improvements over other methods are large too, producing a decrease rate close to or more than 50\%.
It indicates that EMQW produces rather good intervals of 90\%,  80\%,  70\%,  and 60\% probabilities.
From another perspective, QRW also places more weights on the tail-side quantiles, however, its performance is far worse than EMQW. QRW assumes no distribution structure and has great flexibility, verifying that the over-flexibility is harmful and the balance between distribution structure and the flexibility in EMQW is beneficial and necessary.

\subsection{Performance of Ensemble Tree Models}
\begin{table*}[t]
  \centering
  \caption{EICE / EIS results of EMQW and the three ensemble tree models. $^*$ is the best performance among the four. For EICE metric, EMQW wins the best on 16 datasets out of 20. For EIS metric, EMQW wins the best on 9 datasets out of 20.}
  \begin{tabular}{|c|c|c|c|c|}
  \Xhline{1pt}
Dataset $\backslash$ Method & EMQW & \makecell{GBDT + \\Quantile Loss} & \makecell{LightGBM + \\Quantile Loss} & \makecell{Quantile \\Regression Forest} \\ \Xhline{1pt}
Grid & $^*$3.33 / $^*$20.94 & 6.34 / 39.10 & 7.75 / 34.07 & 11.29 / 72.67 \\ \hline
Naval & $^*$2.87 / $^*$10.26 & 5.38 / 39.85 & 4.73 / 24.15 & 11.59 / 36.82 \\ \hline
Appliance & $^*$1.99 / 88.67 & 2.87 / 94.29 & 4.18 / $^*$80.60 & 11.17 / 93.34 \\ \hline
Election & $^*$22.21 / 8.66 & 40.00 / 9.57 & 39.43 / 10.27 & 49.96 / $^*$0.17 \\ \hline
Steel & 8.81 / $^*$1.51 & 4.29 / 8.61 & $^*$4.01 / 5.95 & 14.55 / 5.57 \\ \hline
Facebook1 & $^*$3.03 / 24.73 & 29.00 / 18.39 & 4.75 / $^*$18.05 & 32.02 / 21.26 \\ \hline
Facebook2 & $^*$3.74 / 23.26 & 28.65 / 17.81 & 5.31 / $^*$17.29 & 31.71 / 20.12 \\ \hline
PM2.5 & $^*$1.05 / $^*$54.35 & 1.85 / 81.76 & 2.67 / 74.61 & 5.31 / 81.51 \\ \hline
Bio & $^*$1.10 / $^*$80.51 & 2.21 / 109.07 & 3.02 / 97.76 & 4.58 / 95.48 \\ \hline
Blog & $^*$3.34 / 42.60 & 33.17 / 22.09 & 7.44 / $^*$21.58 & 35.67 / 23.01 \\ \hline
Consumption & 2.05 / 90.98 & $^*$1.65 / 97.71 & 2.27 / 91.50 & 3.92 / $^*$89.93 \\ \hline
Video & $^*$3.85 / 60.72 & 6.99 / 62.29 & 6.99 / 57.72 & 13.27 / $^*$56.19 \\ \hline
GPU & $^*$1.67 / 5.81 & 3.04 / 30.59 & 3.03 / 22.13 & 6.77 / $^*$4.77 \\ \hline
Query & 3.50 / $^*$9.44 & $^*$2.58 / 18.94 & 2.98 / 14.98 & 15.62 / 15.13 \\ \hline
Wave & $^*$3.80 / 12.09 & 4.98 / 4.01 & 5.01 / $^*$2.85 & 14.98 / 6.70 \\ \hline
Air & $^*$1.21 / $^*$29.90 & 1.43 / 32.91 & 1.38 / 30.92 & 3.99 / 36.40 \\ \hline
Year & 1.77 / 106.32 & $^*$1.29 / 107.01 & 1.72 / $^*$104.43 & 2.87 / 117.85 \\ \hline
Relative & $^*$3.86 / $^*$20.56 & 4.06 / 36.43 & 4.89 / 29.60 & 21.42 / 22.11 \\ \hline
Ujiindoorloc & $^*$3.11 / $^*$22.47 & 3.62 / 33.92 & 4.16 / 24.66 & 5.27 / 34.37 \\ \hline
Cup98 & $^*$13.87 / 39.57 & 49.76 / $^*$28.49 & 39.11 / 29.32 & 49.78 / 31.02 \\ \Xhline{1pt}
  \end{tabular}
  \label{tab:interval_ECE_sharpness_boosting}
\end{table*}

Because the proposed EMQW is a type of ensemble (boosting) model, it will be interesting and helpful to compare it to traditional ensemble tree models such as GBDT \cite{friedman2001greedy} and LightGBM \cite{ke2017lightgbm}, both of which are industry-level machine learning models that are extremely popular and useful. They are thought to be the most suitable models for tabular datasets, such as the datasets used in this paper. In both GBDT and LightGBM, one can specify quantile loss as the loss function and thereby predict quantiles. We run them to predict the 99 conditional quantiles of the levels in the set $\Pi$, with each level being handled separately. A third model for comparison is the quantile regression forest proposed in \cite{meinshausen2006quantile}, which is popular too.

We tune the hyper-parameters of GBDT and LightGBM on the 20\% validation set: the maximum depth of a single tree, and the number of trees. 
For quantile regression forest, the fraction of features used in the node splitting is tuned as well.
For each $\tau\in\Pi$, this tuning will be done separately. We report their results of EICE and EIS together in Table \ref{tab:interval_ECE_sharpness_boosting}. From the table, we find that for EICE, our EMQW wins the best on 16 datasets out of 20. For EIS, EMQW wins the best on 9 datasets out of 20. Conclusively, EMQW shows better performance than these traditional models do. 

\subsection{Interpretation of Learning Results}

\begin{figure*}[t]
\begin{center}
\begin{minipage}[t]{.16\linewidth}
\centerline{\includegraphics[width=\linewidth]{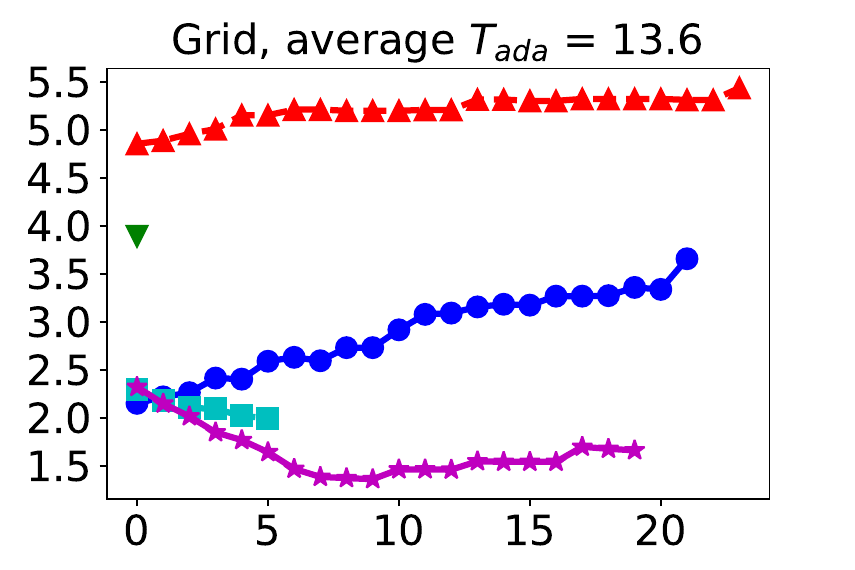}}
\end{minipage}
\begin{minipage}[t]{.16\linewidth}
\centerline{\includegraphics[width=\linewidth]{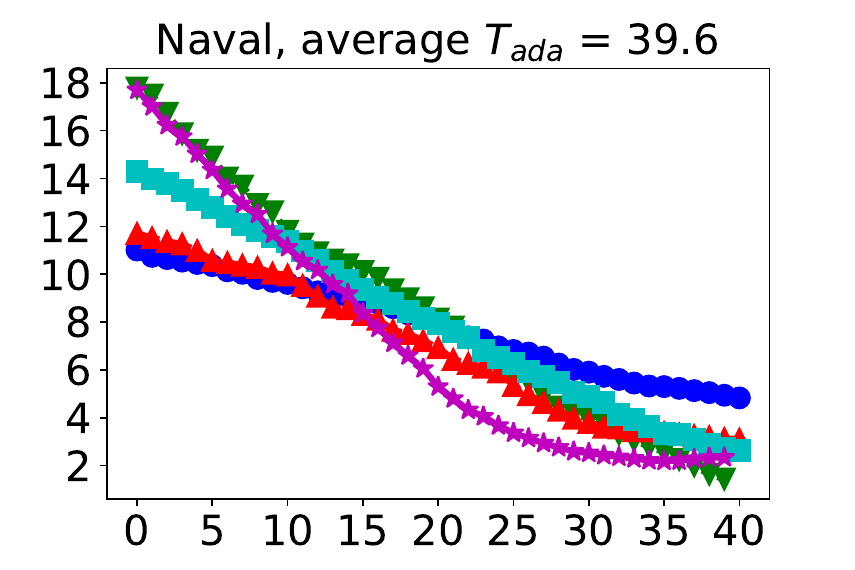}}
\end{minipage}
\begin{minipage}[t]{.16\linewidth}
\centerline{\includegraphics[width=\linewidth]{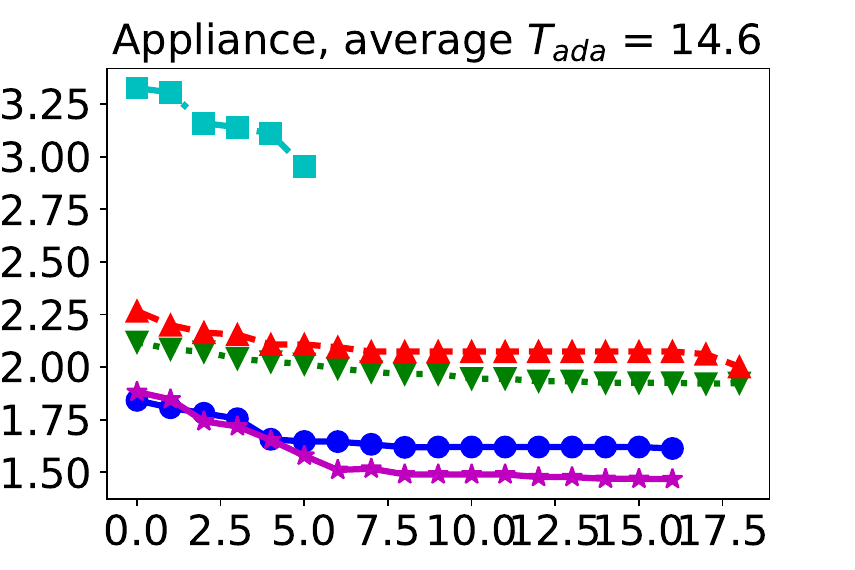}}
\end{minipage}
\begin{minipage}[t]{.16\linewidth}
\centerline{\includegraphics[width=\linewidth]{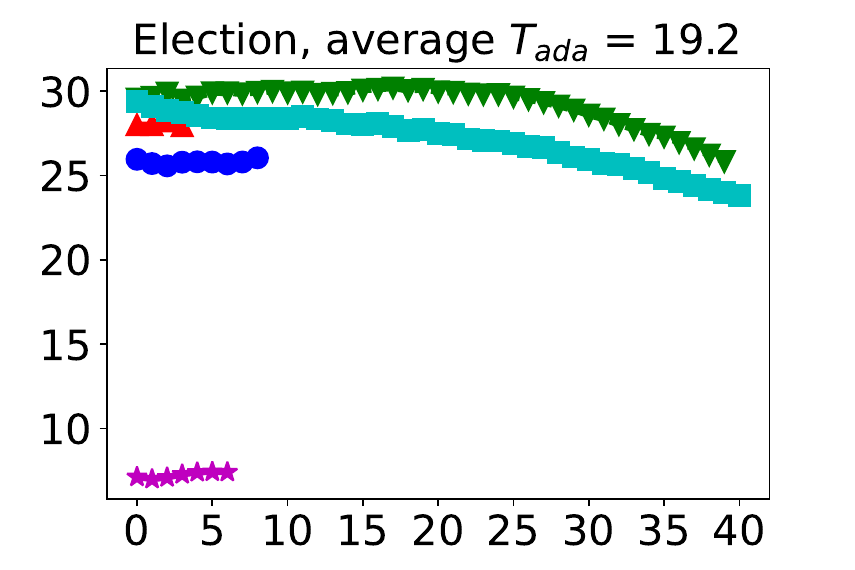}}
\end{minipage}
\begin{minipage}[t]{.16\linewidth}
\centerline{\includegraphics[width=\linewidth]{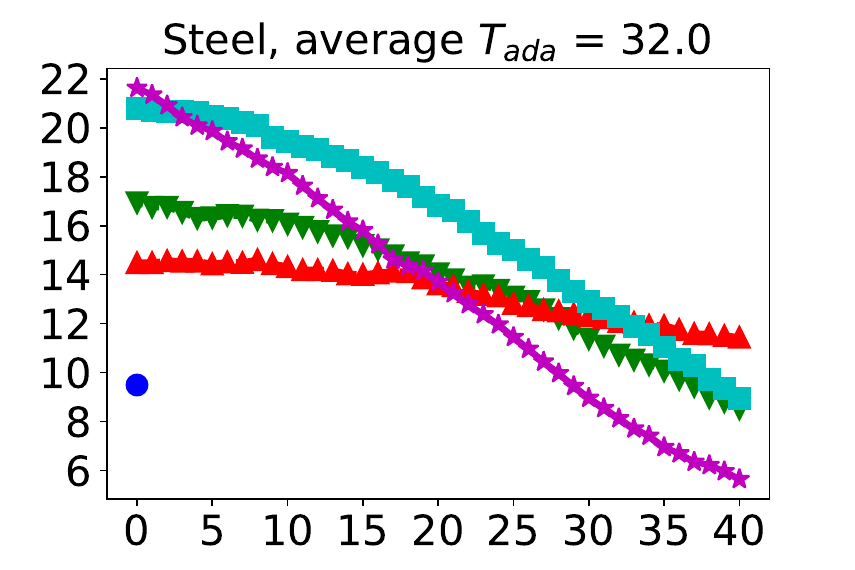}}
\end{minipage}\\
\begin{minipage}[t]{.16\linewidth}
\centerline{\includegraphics[width=\linewidth]{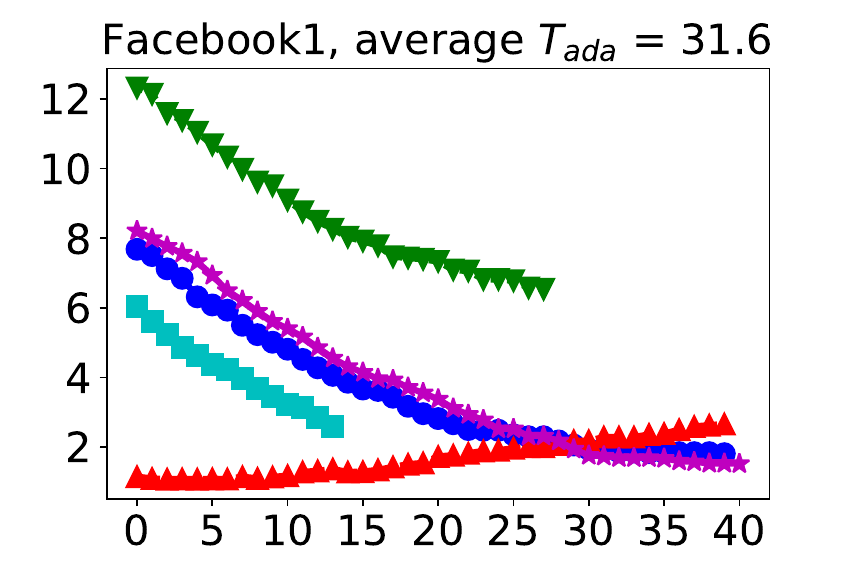}}
\end{minipage}
\begin{minipage}[t]{.16\linewidth}
\centerline{\includegraphics[width=\linewidth]{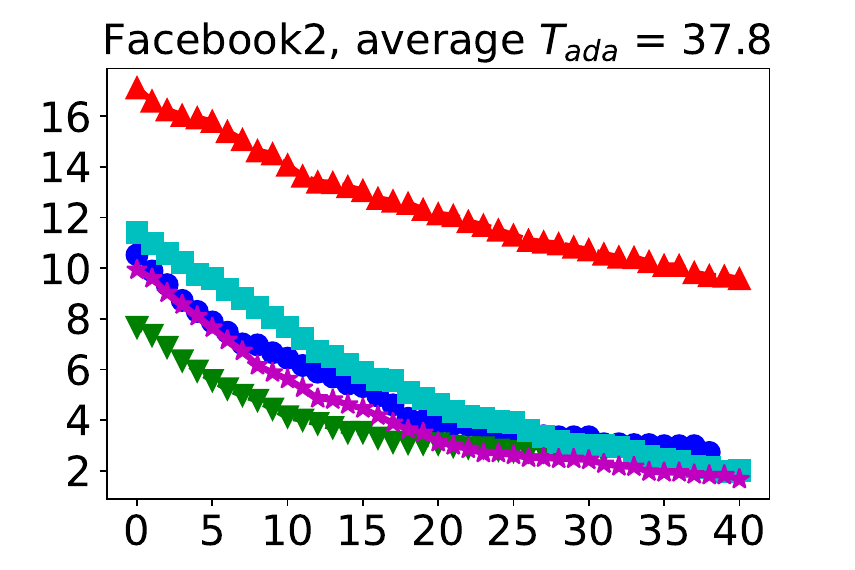}}
\end{minipage}
\begin{minipage}[t]{.16\linewidth}
\centerline{\includegraphics[width=\linewidth]{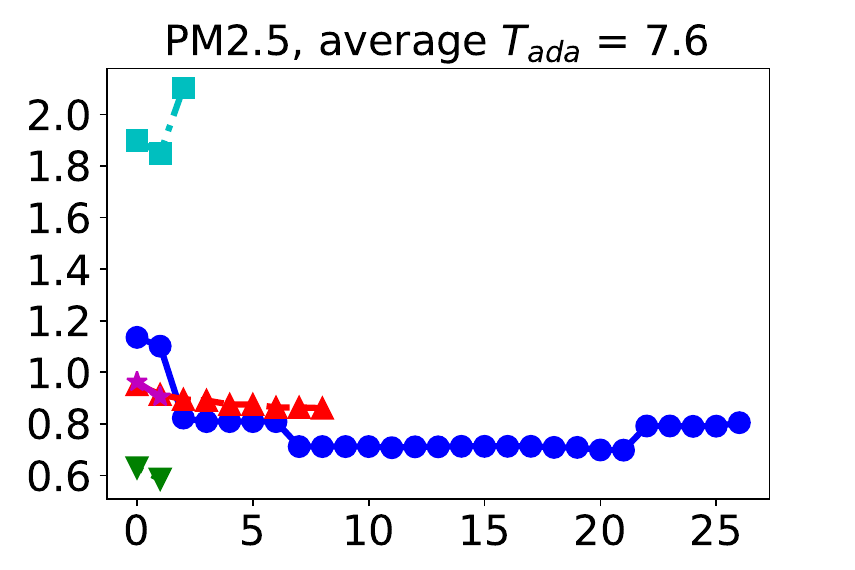}}
\end{minipage}
\begin{minipage}[t]{.16\linewidth}
\centerline{\includegraphics[width=\linewidth]{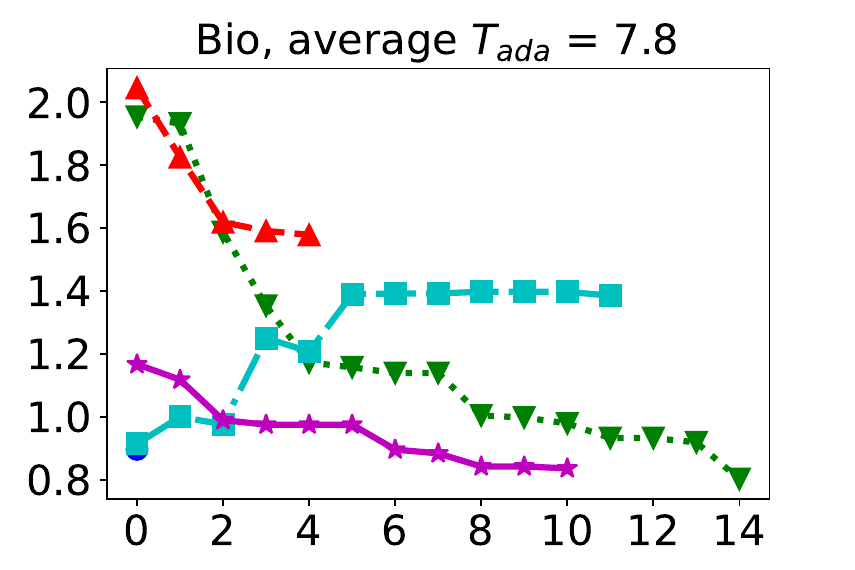}}
\end{minipage}
\begin{minipage}[t]{.16\linewidth}
\centerline{\includegraphics[width=\linewidth]{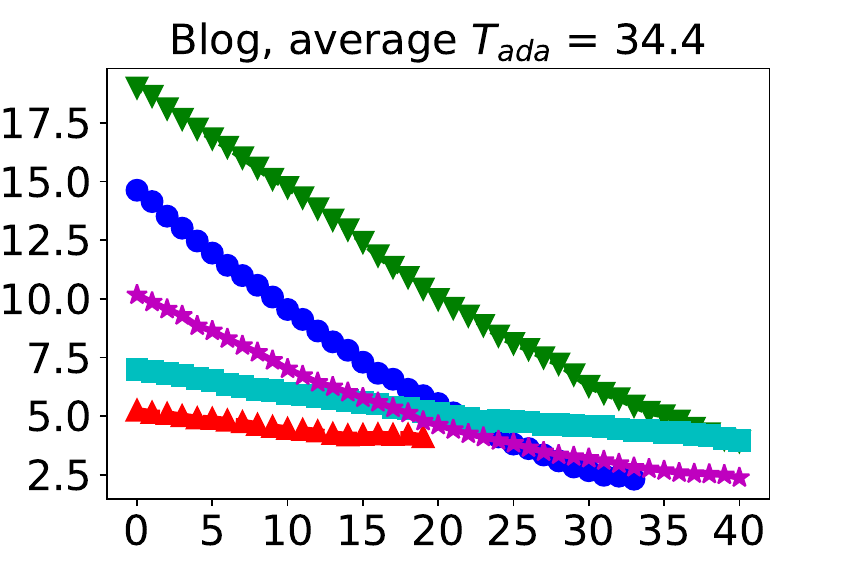}}
\end{minipage}\\
\begin{minipage}[t]{.16\linewidth}
\centerline{\includegraphics[width=\linewidth]{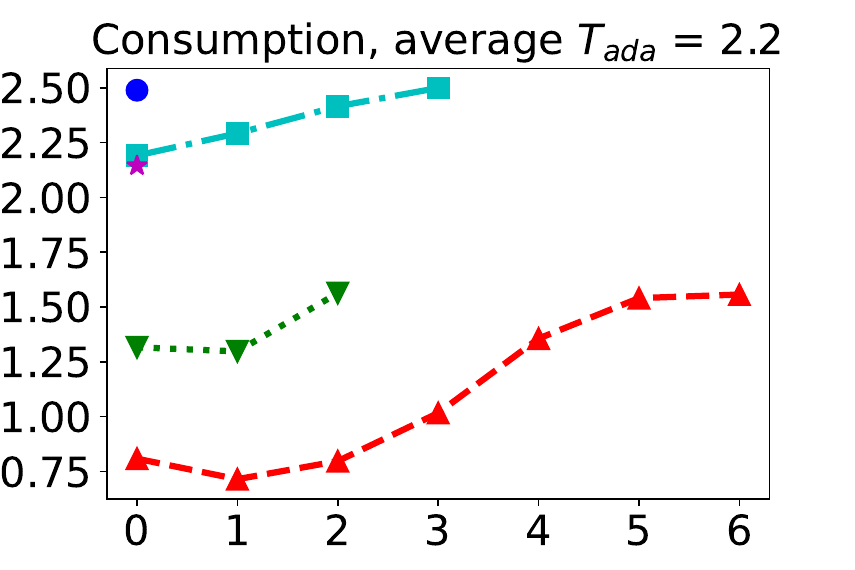}}
\end{minipage}
\begin{minipage}[t]{.16\linewidth}
\centerline{\includegraphics[width=\linewidth]{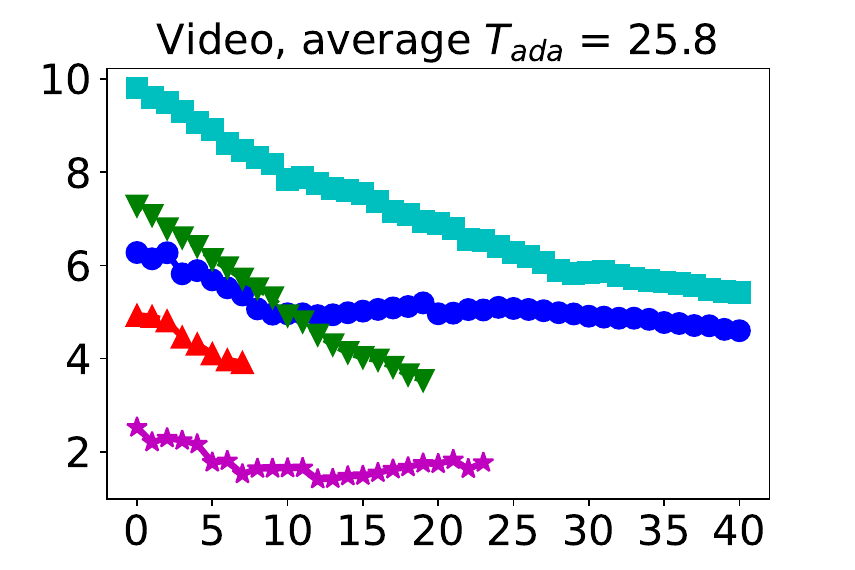}}
\end{minipage}
\begin{minipage}[t]{.16\linewidth}
\centerline{\includegraphics[width=\linewidth]{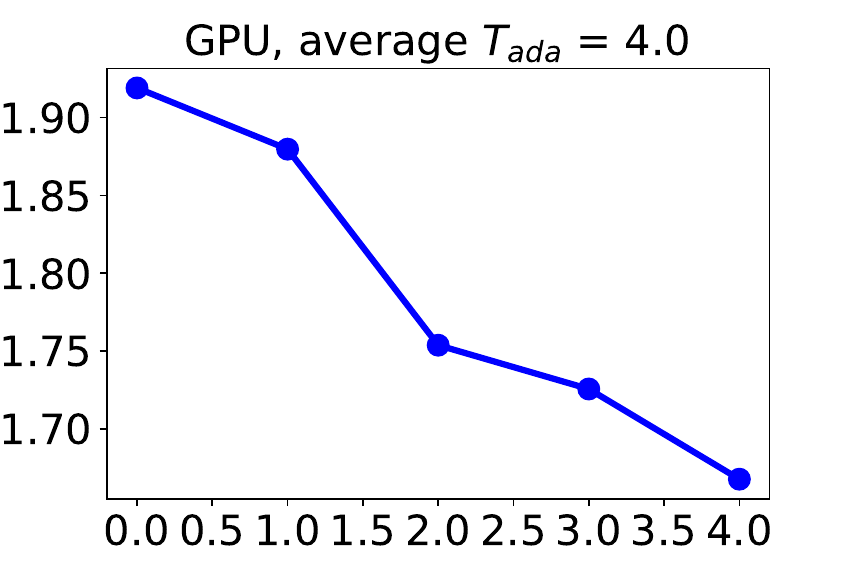}}
\end{minipage}
\begin{minipage}[t]{.16\linewidth}
\centerline{\includegraphics[width=\linewidth]{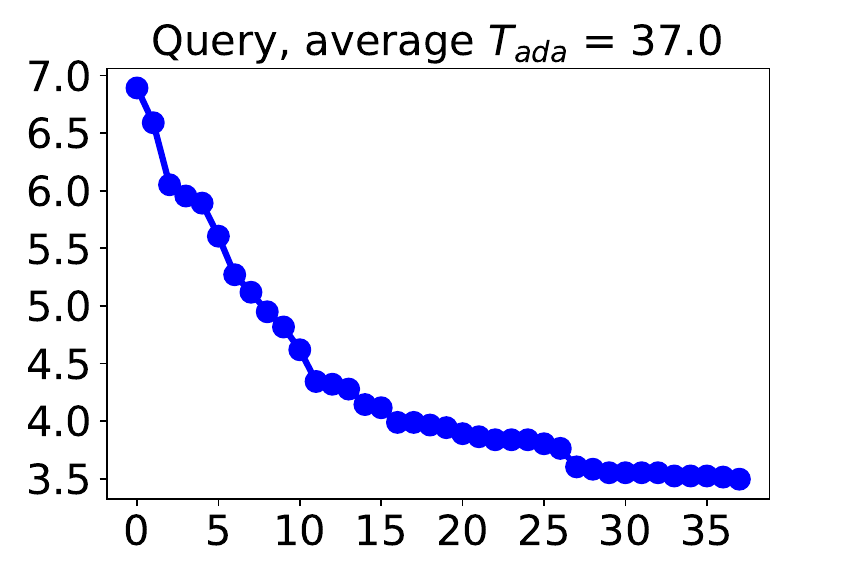}}
\end{minipage}
\begin{minipage}[t]{.16\linewidth}
\centerline{\includegraphics[width=\linewidth]{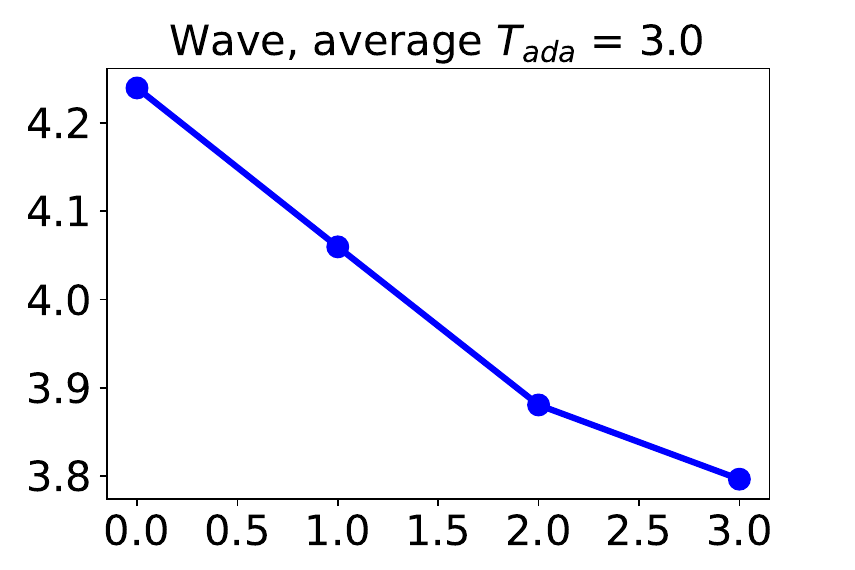}}
\end{minipage}\\
\begin{minipage}[t]{.16\linewidth}
\centerline{\includegraphics[width=\linewidth]{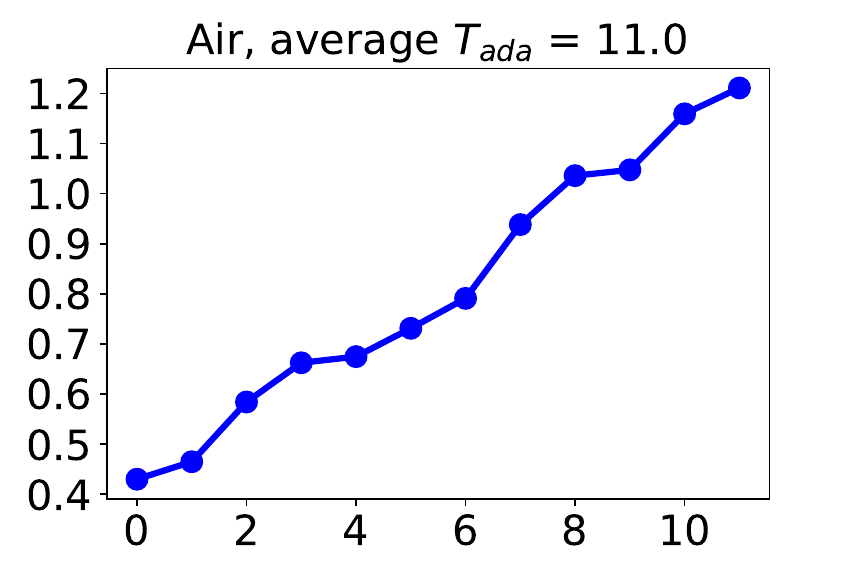}}
\end{minipage}
\begin{minipage}[t]{.16\linewidth}
\centerline{\includegraphics[width=\linewidth]{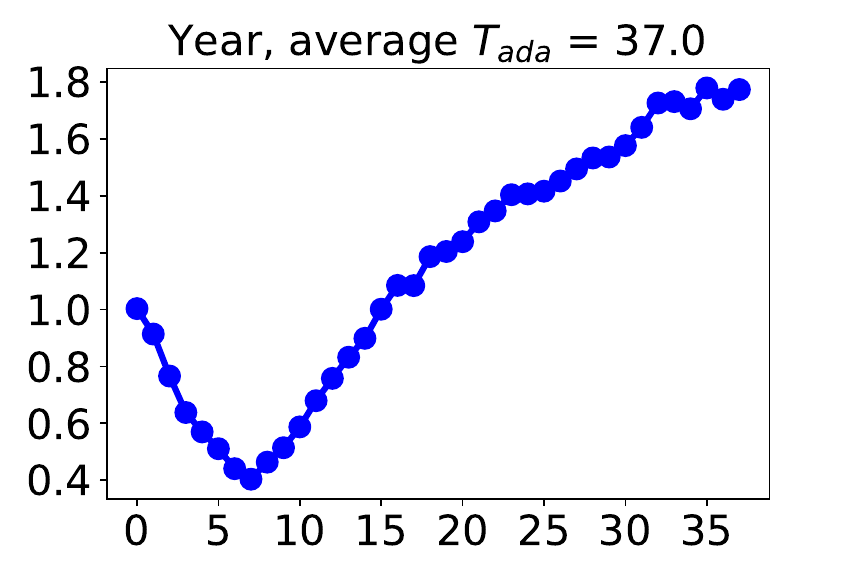}}
\end{minipage}
\begin{minipage}[t]{.16\linewidth}
\centerline{\includegraphics[width=\linewidth]{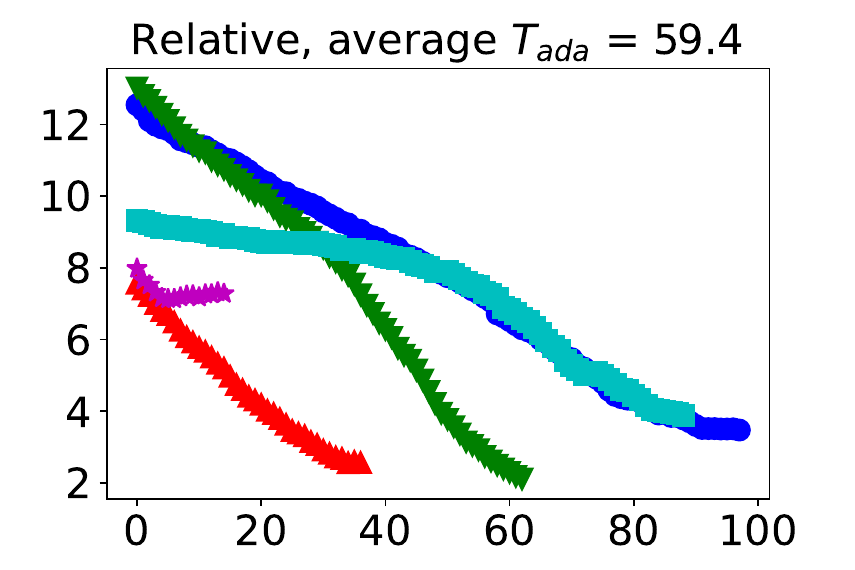}}
\end{minipage}
\begin{minipage}[t]{.16\linewidth}
\centerline{\includegraphics[width=\linewidth]{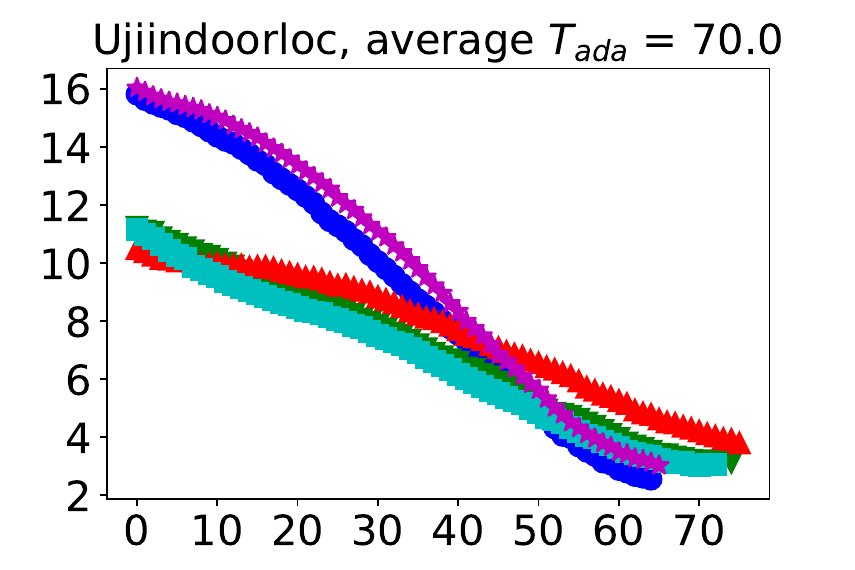}}
\end{minipage}
\begin{minipage}[t]{.16\linewidth}
\centerline{\includegraphics[width=\linewidth]{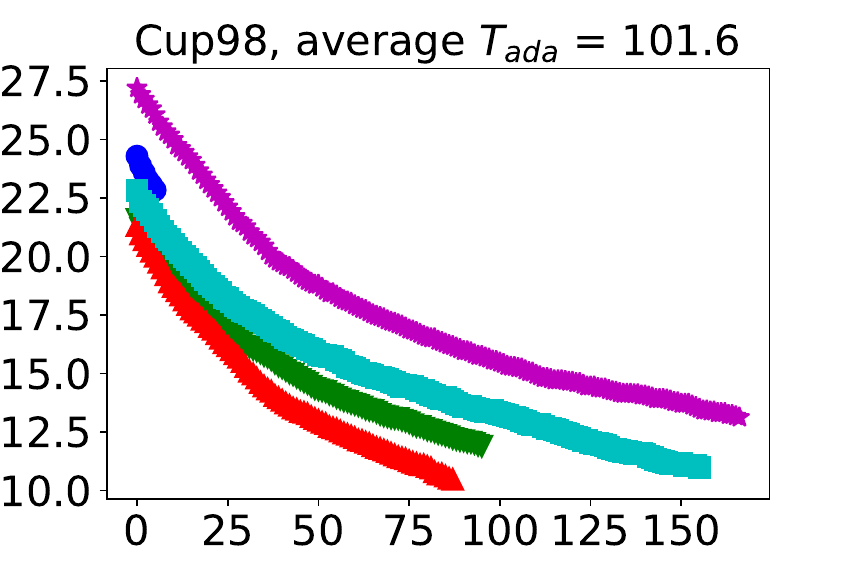}}
\end{minipage}
  \caption{The changing of EICE on the testing set as the step $t$ increases from 0 to $T_{ada}$ in EMQW. Each subfigure corresponds to one dataset, and the lines of different colors and styles within one subfigure represent different training/testing splits. In the vast majority of lines, EICE decreases as $t$ increases with very few exceptions, indicating that the ensemble steps in EMQW work correctly as we expect.}
  \label{fig:T_ada}
  \end{center}
  \vspace{-1em}
  \end{figure*}

\begin{figure*}[t]
\begin{center}
\begin{minipage}[t]{.165\linewidth}
\centerline{\includegraphics[width=\linewidth]{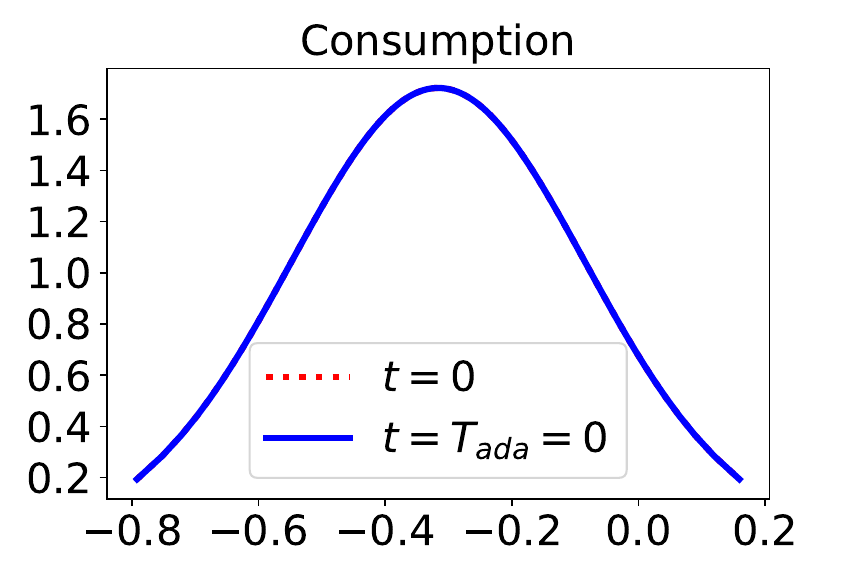}}
\end{minipage}\hfill
\begin{minipage}[t]{.165\linewidth}
\centerline{\includegraphics[width=\linewidth]{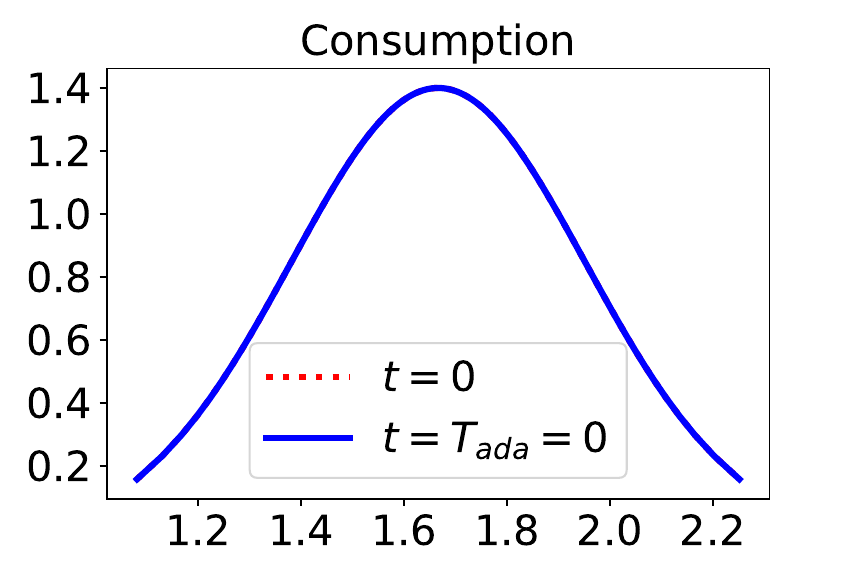}}
\end{minipage}\hfill
\begin{minipage}[t]{.165\linewidth}
\centerline{\includegraphics[width=\linewidth]{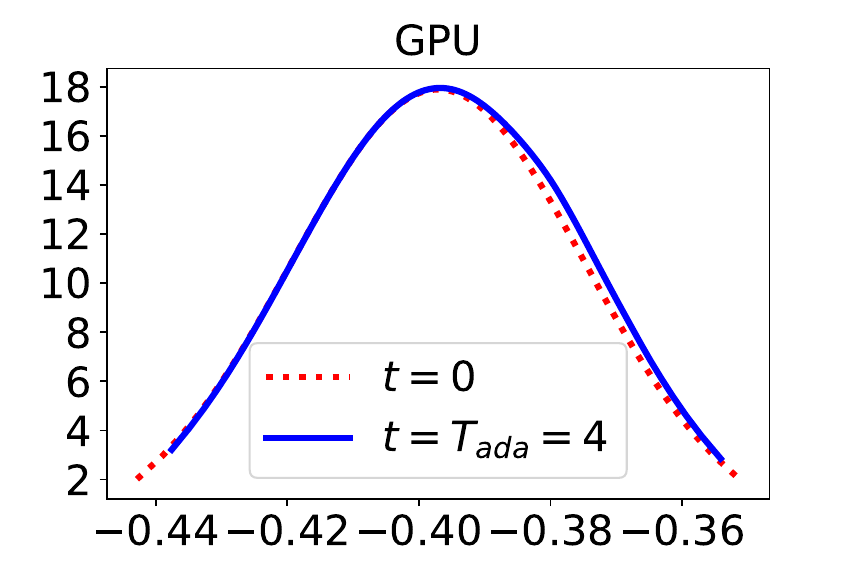}}
\end{minipage}\hfill
\begin{minipage}[t]{.165\linewidth}
\centerline{\includegraphics[width=\linewidth]{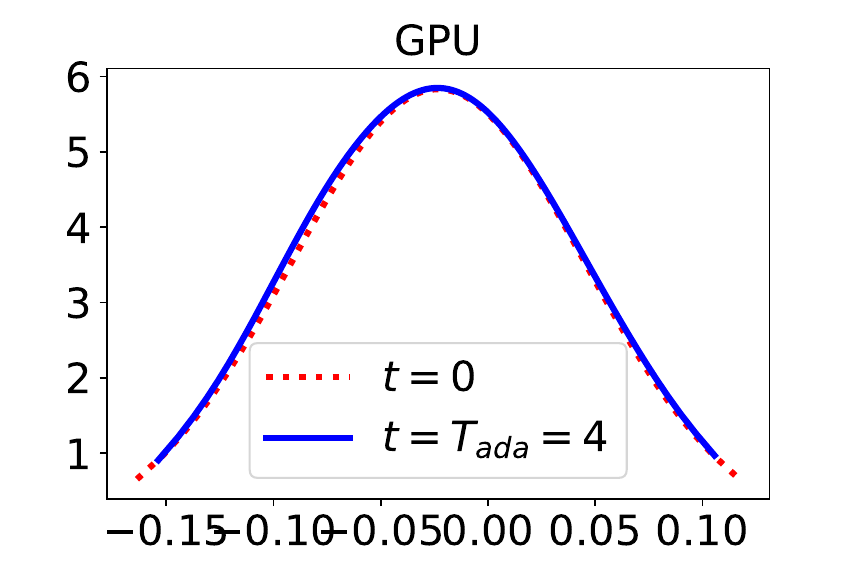}}
\end{minipage}\hfill
\begin{minipage}[t]{.165\linewidth}
\centerline{\includegraphics[width=\linewidth]{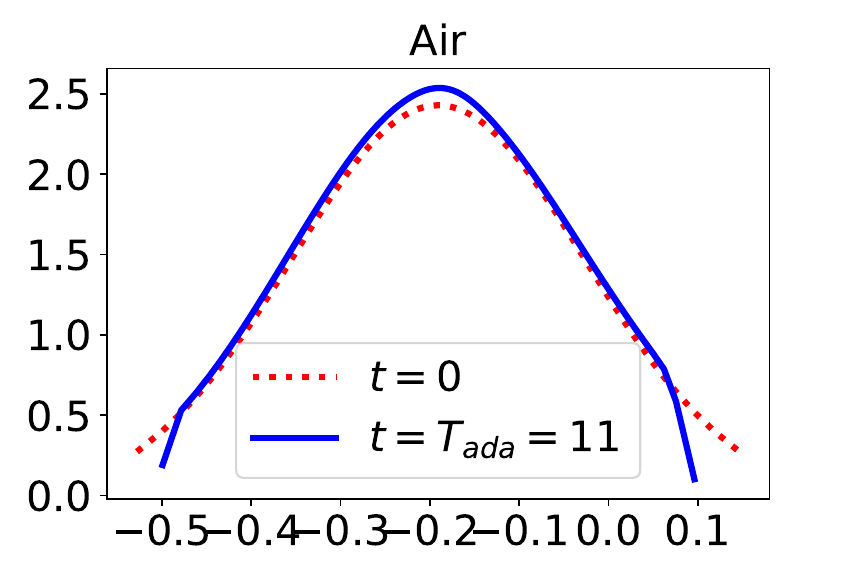}}
\end{minipage}\hfill
\begin{minipage}[t]{.165\linewidth}
\centerline{\includegraphics[width=\linewidth]{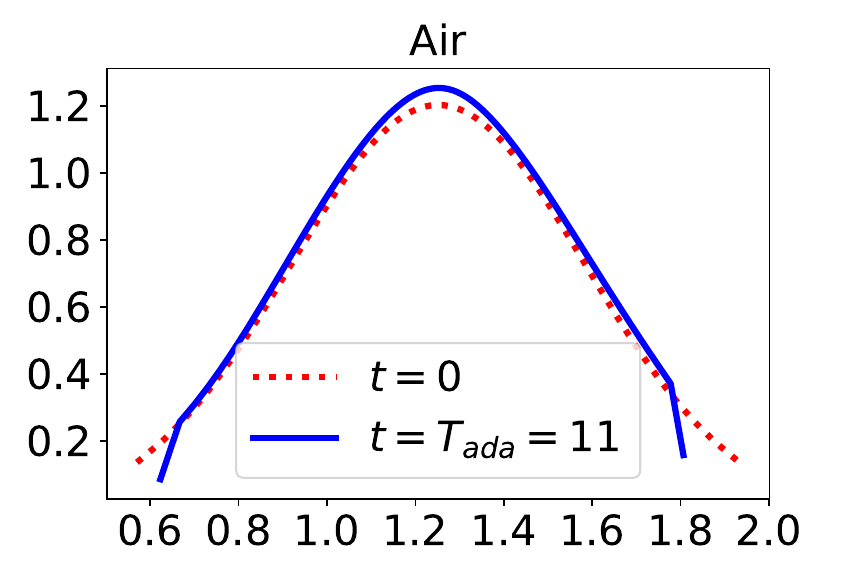}}
\end{minipage}\\
\begin{minipage}[t]{.165\linewidth}
\centerline{\includegraphics[width=\linewidth]{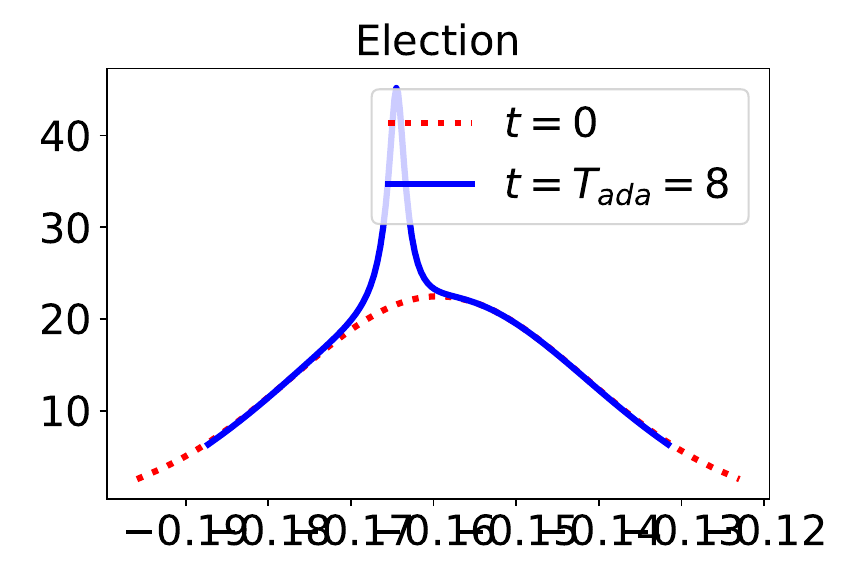}}
\end{minipage}\hfill
\begin{minipage}[t]{.165\linewidth}
\centerline{\includegraphics[width=\linewidth]{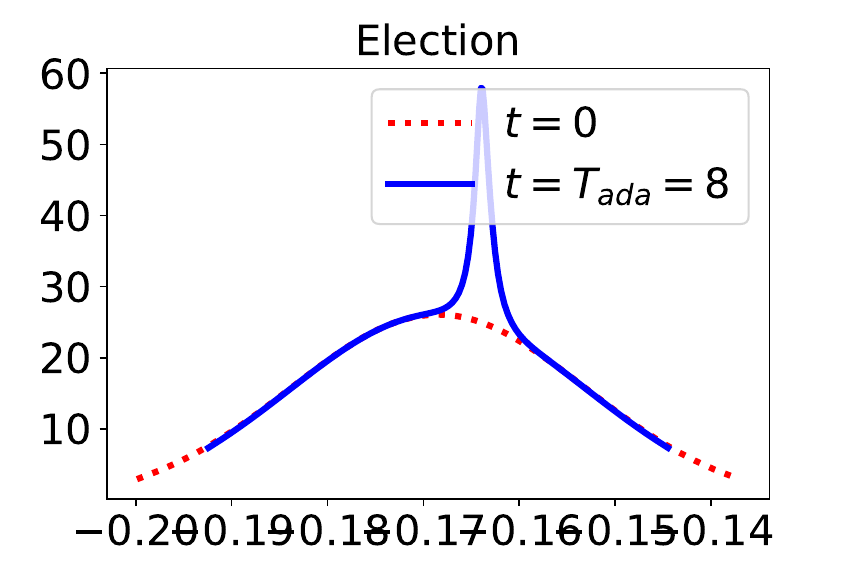}}
\end{minipage}\hfill
\begin{minipage}[t]{.165\linewidth}
\centerline{\includegraphics[width=\linewidth]{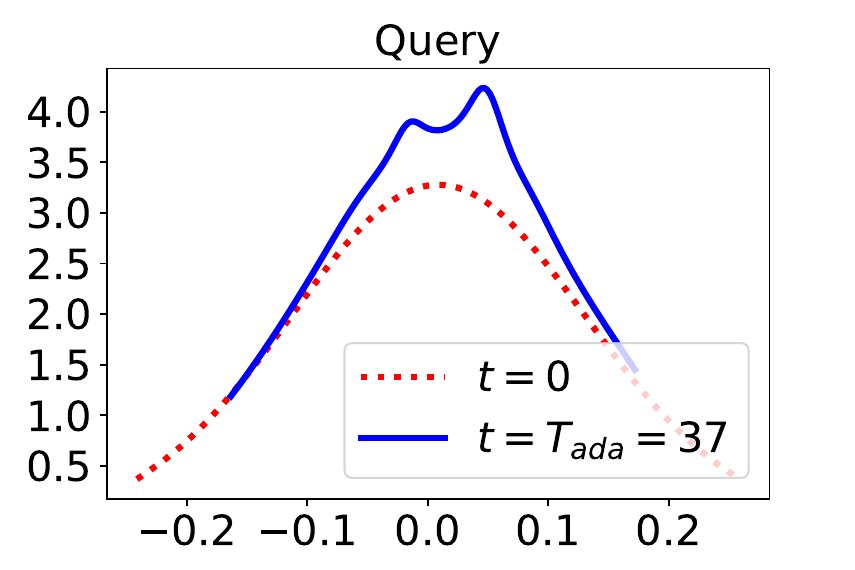}}
\end{minipage}\hfill
\begin{minipage}[t]{.165\linewidth}
\centerline{\includegraphics[width=\linewidth]{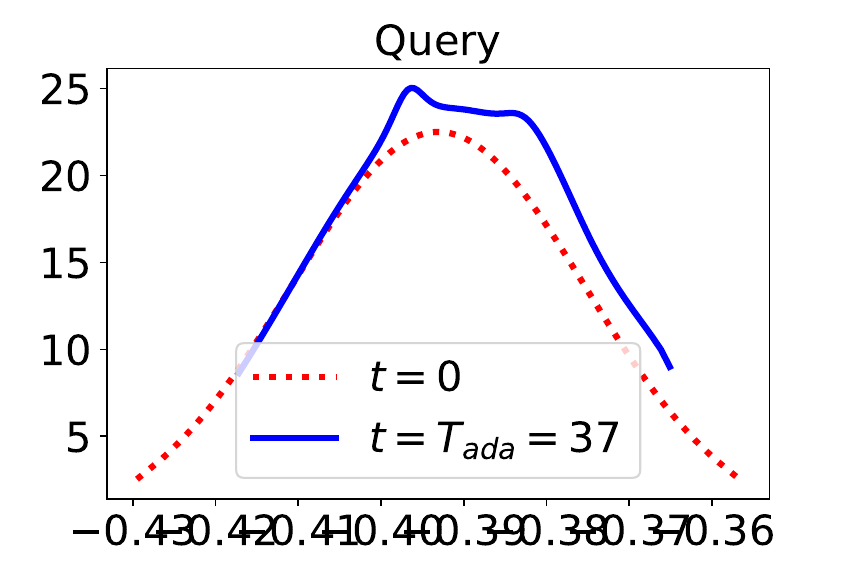}}
\end{minipage}\hfill
\begin{minipage}[t]{.165\linewidth}
\centerline{\includegraphics[width=\linewidth]{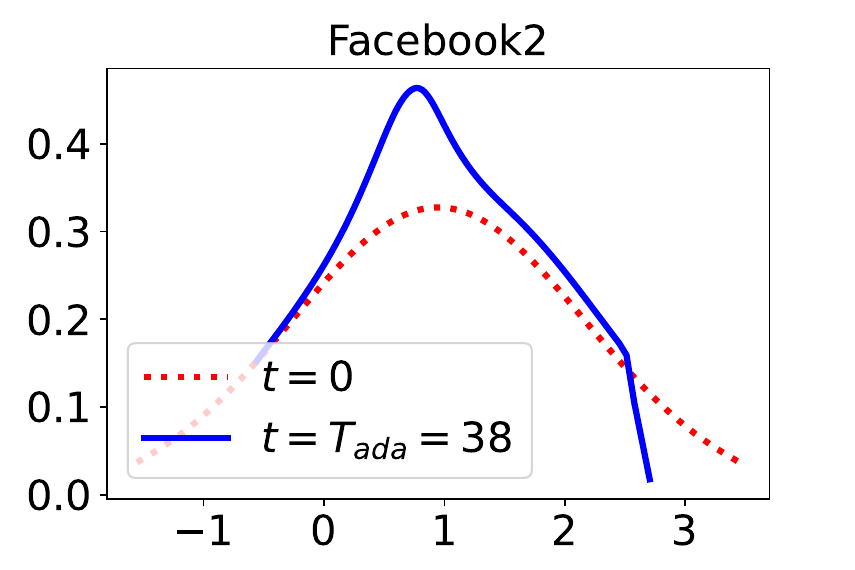}}
\end{minipage}\hfill
\begin{minipage}[t]{.165\linewidth}
\centerline{\includegraphics[width=\linewidth]{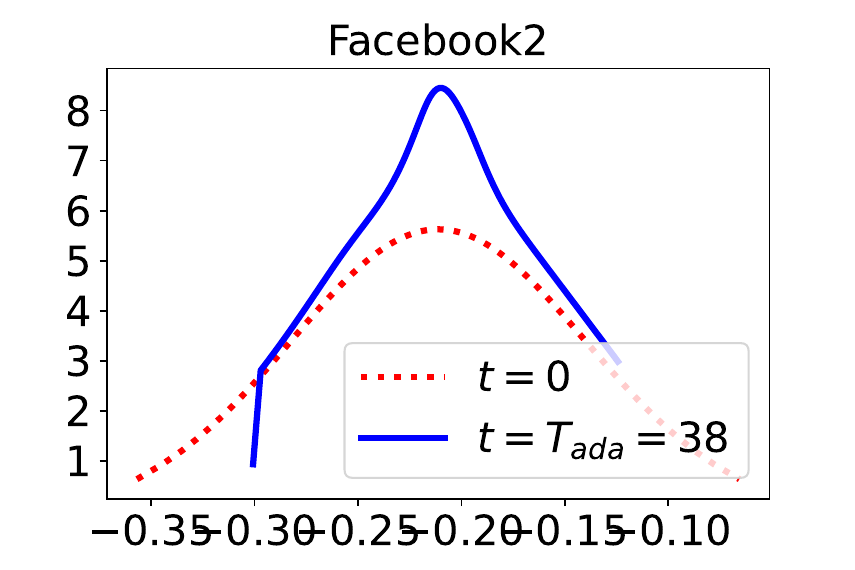}}
\end{minipage}\\
\begin{minipage}[t]{.165\linewidth}
\centerline{\includegraphics[width=\linewidth]{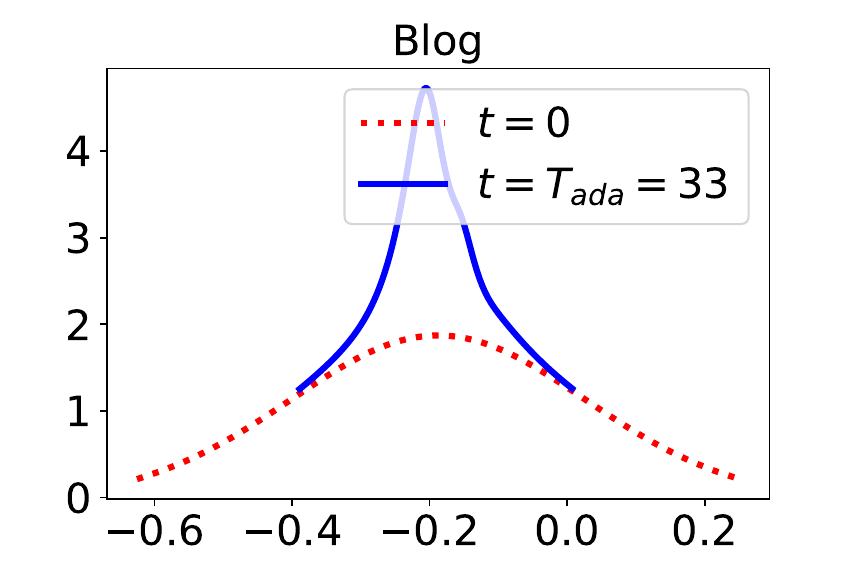}}
\end{minipage}\hfill
\begin{minipage}[t]{.165\linewidth}
\centerline{\includegraphics[width=\linewidth]{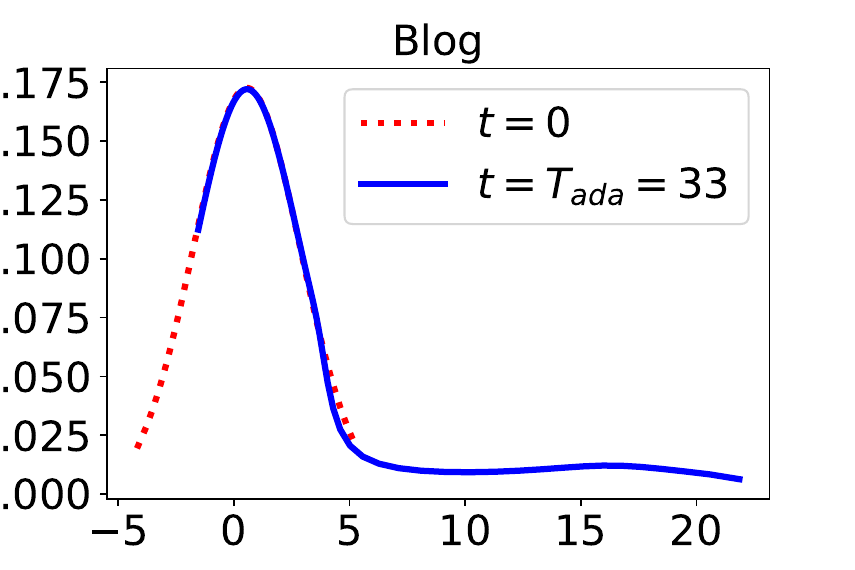}}
\end{minipage}\hfill
\begin{minipage}[t]{.165\linewidth}
\centerline{\includegraphics[width=\linewidth]{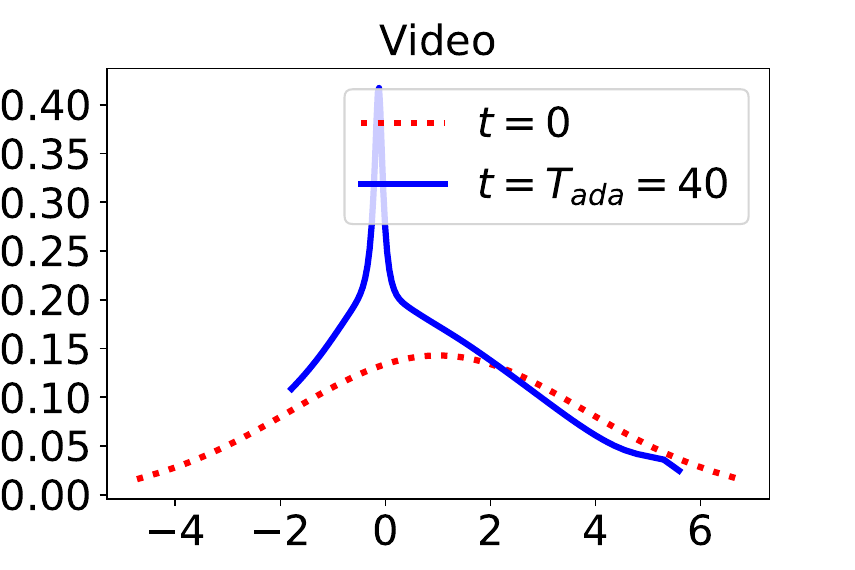}}
\end{minipage}\hfill
\begin{minipage}[t]{.165\linewidth}
\centerline{\includegraphics[width=\linewidth]{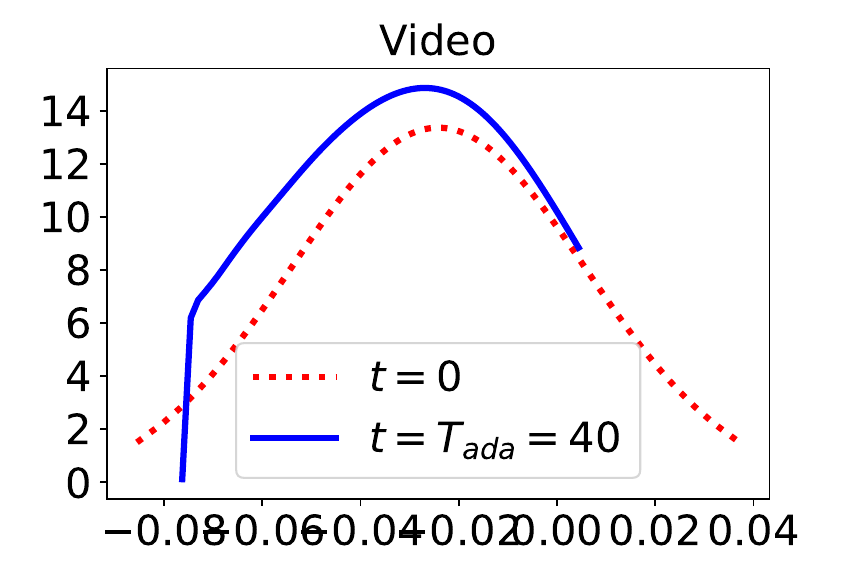}}
\end{minipage}\hfill
\begin{minipage}[t]{.165\linewidth}
\centerline{\includegraphics[width=\linewidth]{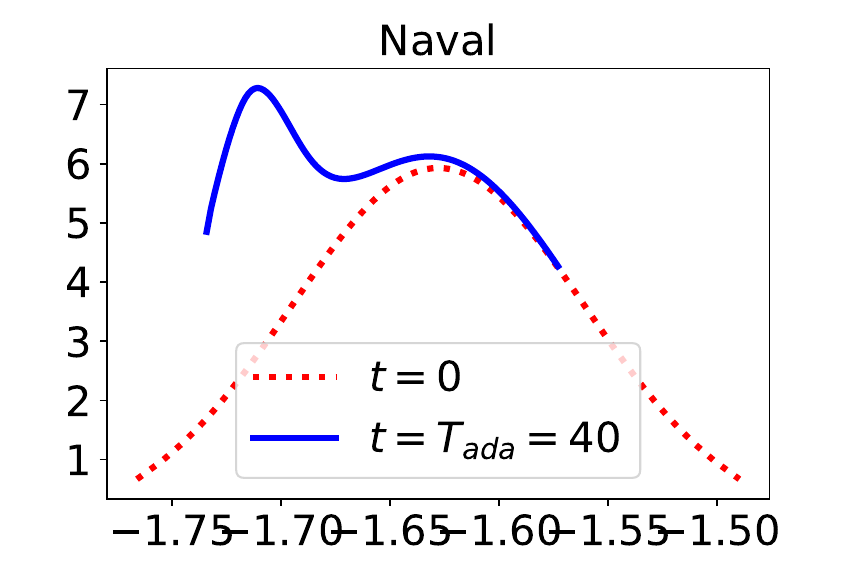}}
\end{minipage}\hfill
\begin{minipage}[t]{.165\linewidth}
\centerline{\includegraphics[width=\linewidth]{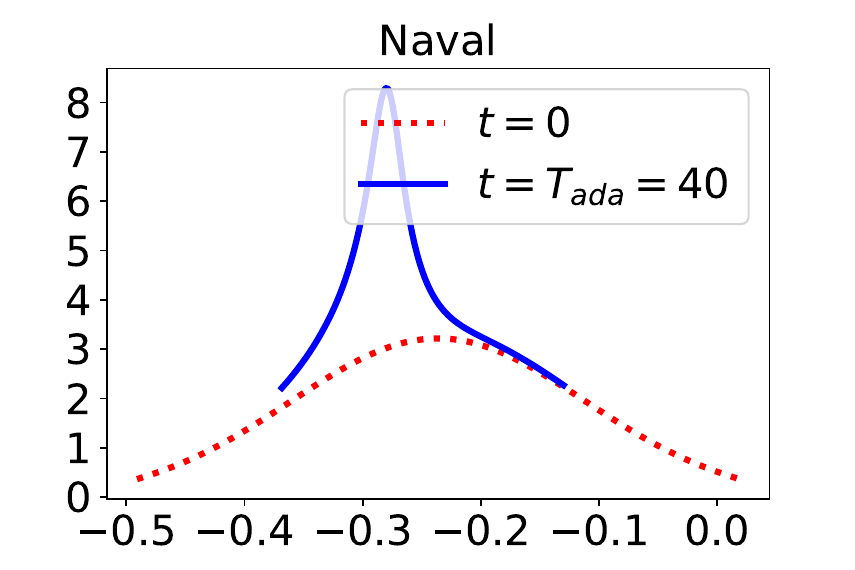}}
\end{minipage}
  \caption{Densities implied from the 99 conditional quantiles predicted by EMQW. For each dataset, we select two data points from the testing set and plot their densities of $\mathbb{P}(\mathbf{y}|\mathbf{X}=x)$. Red lines are the results at $t=0$ and blue ones are the results at $t=T_{ada}$. The first row of subfigures: EMQW produces exactly Gaussian, or approximately Gaussian distributions with slight differences. The second row: the densities show various peaks and asymmetric shapes. The third row: the densities show more diverse shapes, such as a long right tail, a left bound, and a multi-peak shape. The variety may occur within one dataset, revealing the complexity of real data and the ability of EMQW.}
  \label{fig:density}
  \end{center}
  \vspace{-1em}
  \end{figure*}

\begin{figure}[t]
\begin{center}
\begin{minipage}[t]{.33\linewidth}
\centerline{\includegraphics[width=\linewidth]{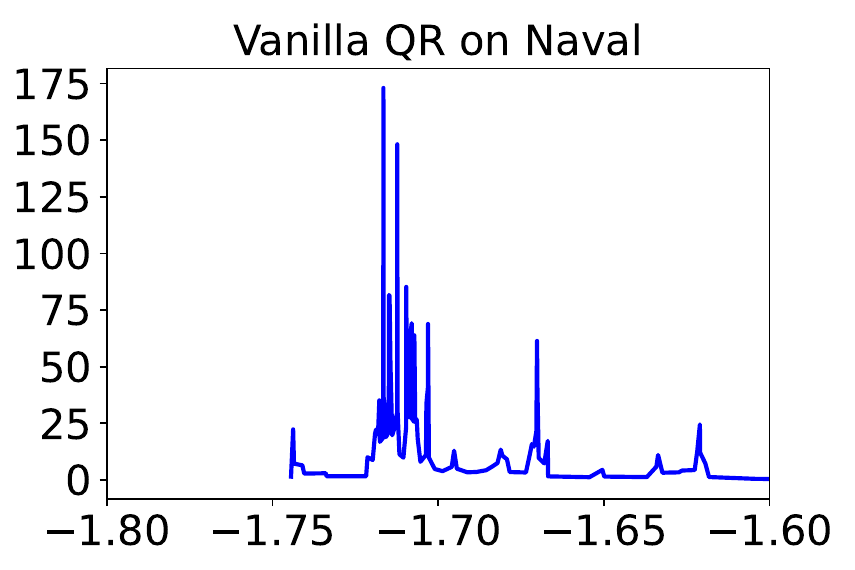}}
\end{minipage}\hfill
\begin{minipage}[t]{.33\linewidth}
\centerline{\includegraphics[width=\linewidth]{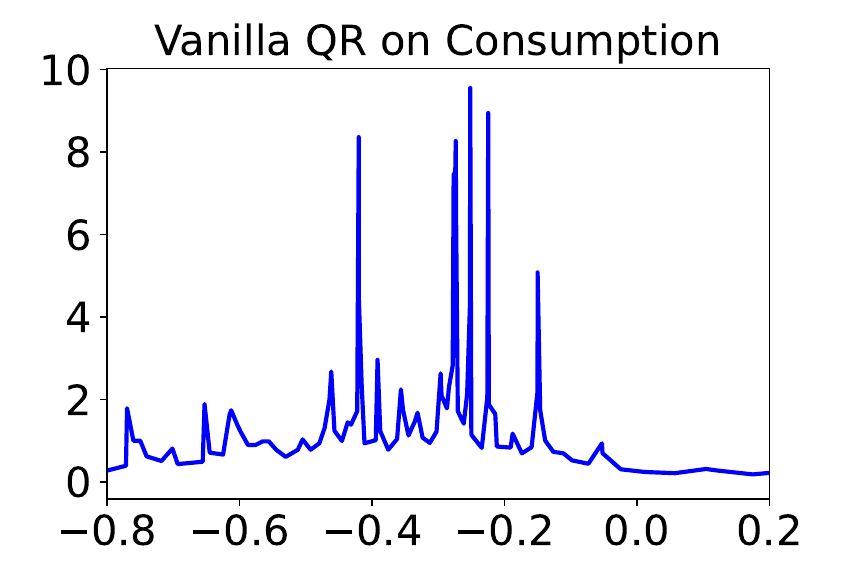}}
\end{minipage}\hfill
\begin{minipage}[t]{.33\linewidth}
\centerline{\includegraphics[width=\linewidth]{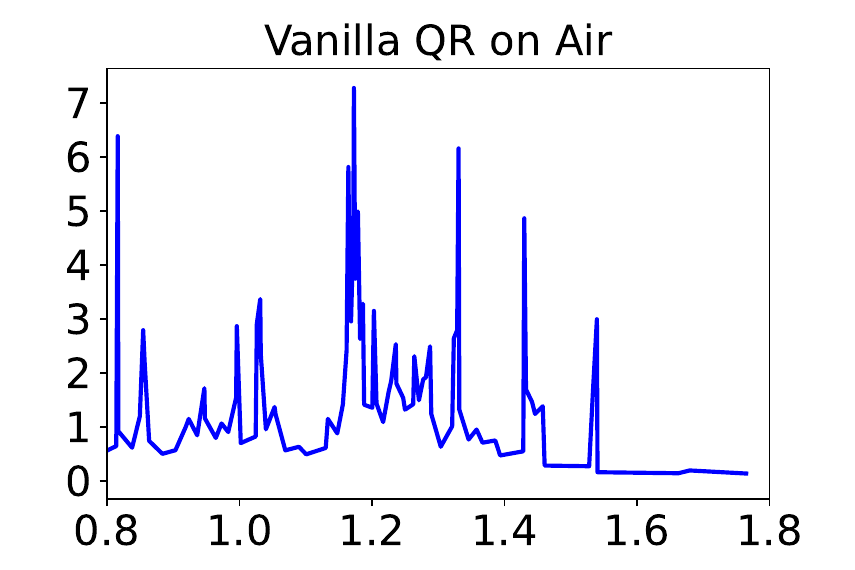}}
\end{minipage}\\
\begin{minipage}[t]{.33\linewidth}
\centerline{\includegraphics[width=\linewidth]{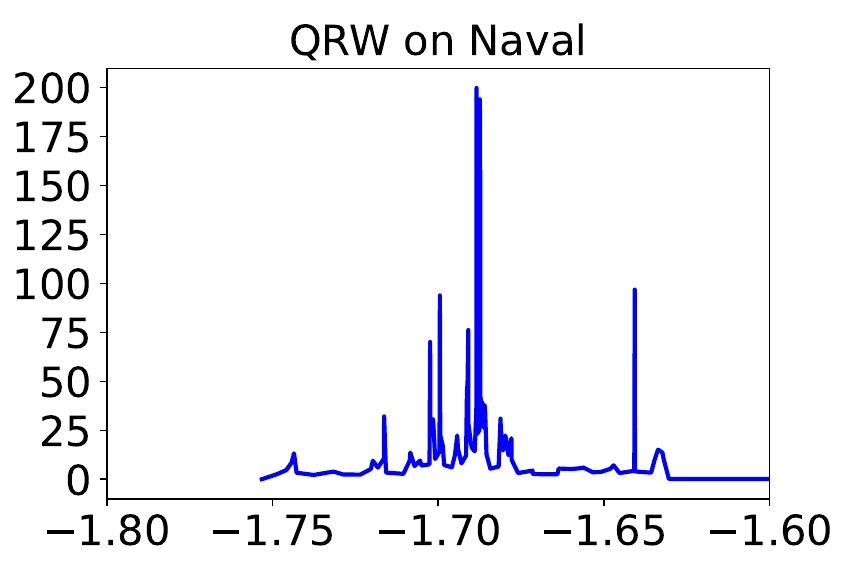}}
\end{minipage}\hfill
\begin{minipage}[t]{.33\linewidth}
\centerline{\includegraphics[width=\linewidth]{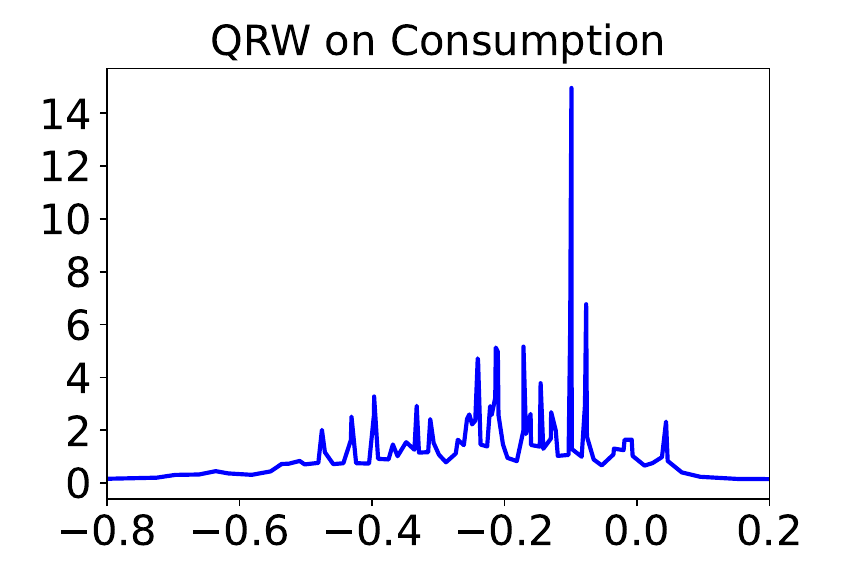}}
\end{minipage}\hfill
\begin{minipage}[t]{.33\linewidth}
\centerline{\includegraphics[width=\linewidth]{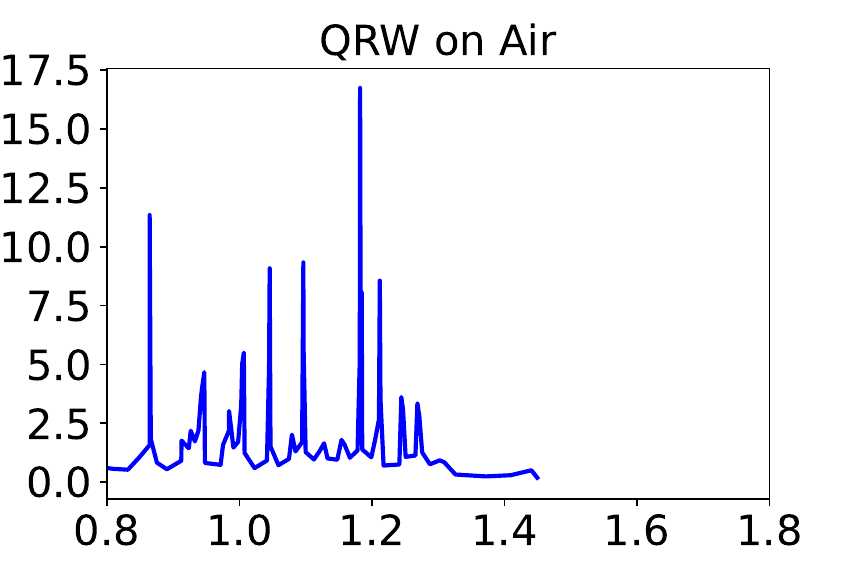}}
\end{minipage}\\
\begin{minipage}[t]{.33\linewidth}
\centerline{\includegraphics[width=\linewidth]{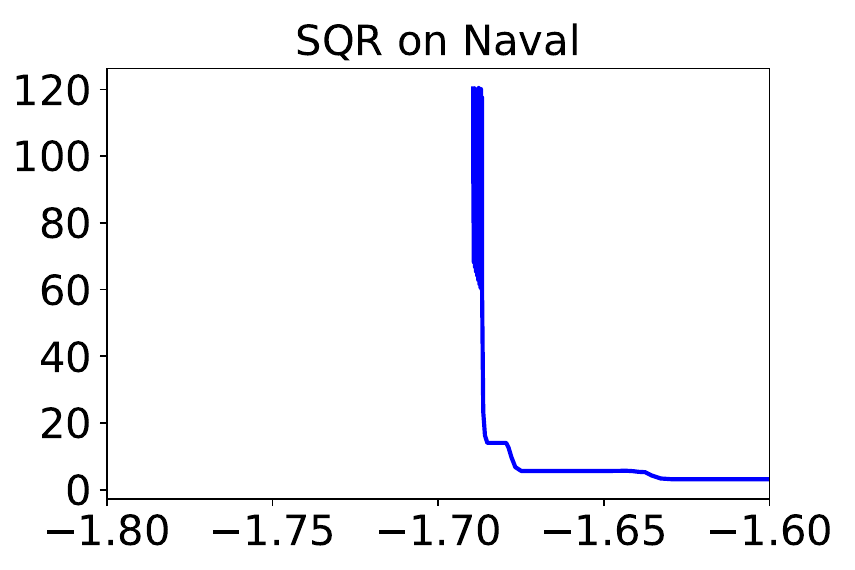}}
\end{minipage}\hfill
\begin{minipage}[t]{.33\linewidth}
\centerline{\includegraphics[width=\linewidth]{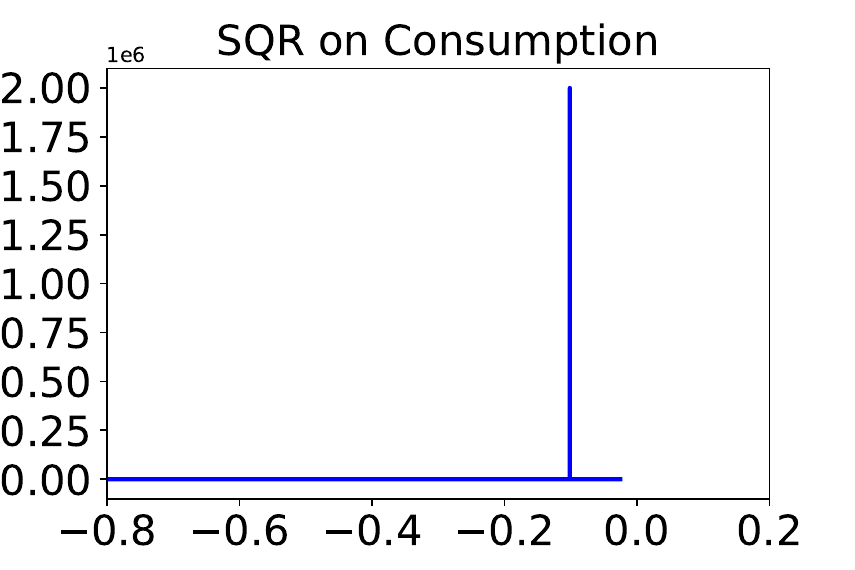}}
\end{minipage}\hfill
\begin{minipage}[t]{.33\linewidth}
\centerline{\includegraphics[width=\linewidth]{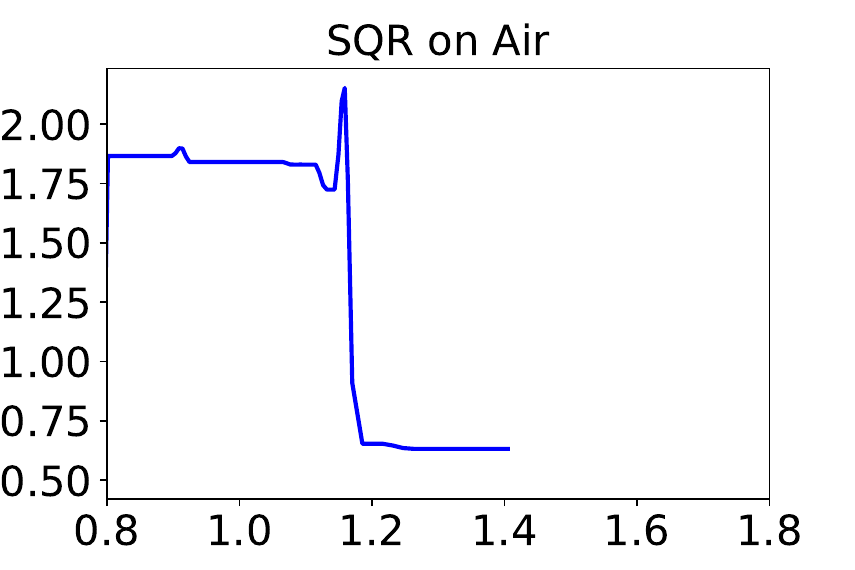}}
\end{minipage}\\
\begin{minipage}[t]{.33\linewidth}
\centerline{\includegraphics[width=\linewidth]{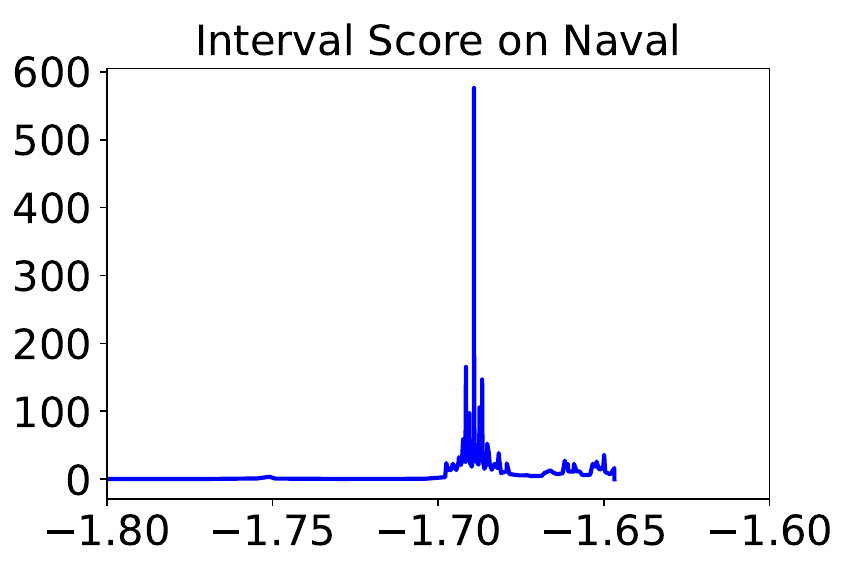}}
\end{minipage}\hfill
\begin{minipage}[t]{.33\linewidth}
\centerline{\includegraphics[width=\linewidth]{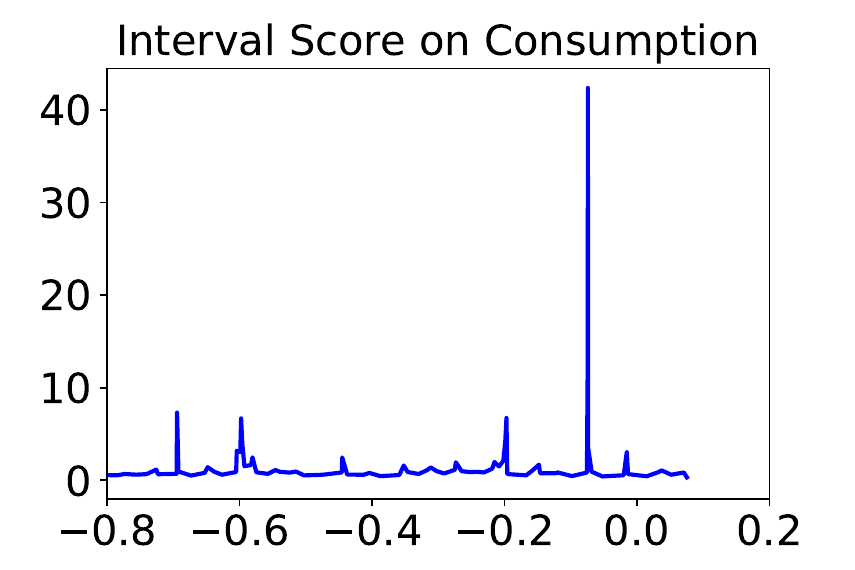}}
\end{minipage}\hfill
\begin{minipage}[t]{.33\linewidth}
\centerline{\includegraphics[width=\linewidth]{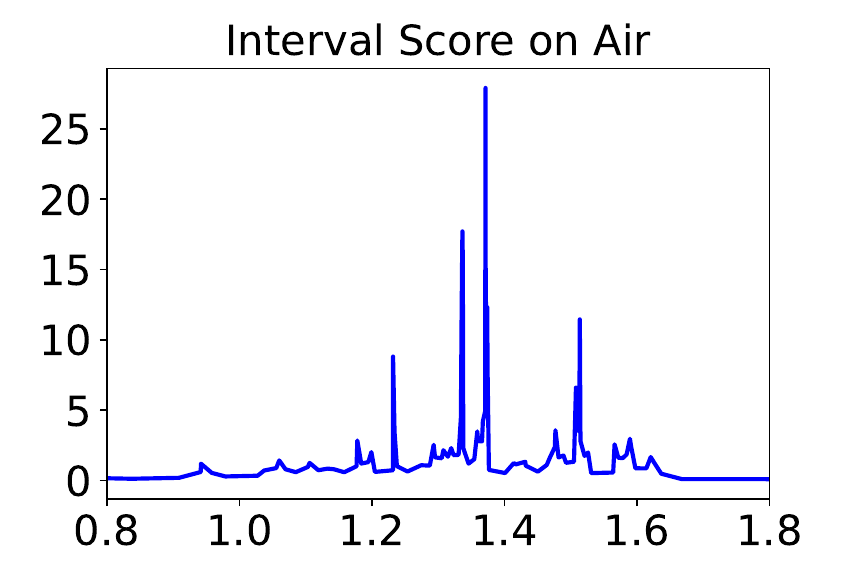}}
\end{minipage}
 \caption{Densities implied from the 99 conditional quantiles predicted by four quantile-based methods: Vanilla QR, QRW, SQR, and Interval Score. Variance networks and Dropout methods all produce Gaussian densities, so unnecessary to plot. We select one data point from the testing set of the dataset: Naval, Consumption, and Air, and plot its densities of $\mathbb{P}(\mathbf{y}|\mathbf{X}=x)$ given by the four methods. We can see that all of them produce rough or even invalid densities.}
  \label{fig:other_density}
  \end{center}
  \vspace{-1em}
  \end{figure}

We visualize the learning process and some learning results of EMQW for better interpretation. This helps us understand how EMQW works and why it gives superior performance.

\textbf{Performance variation of EMQW with step $t$}. We are interested in the performance variation of EMQW on the testing set with respect to $t$. We plot in Fig. \ref{fig:T_ada} the varying performance of the metric EICE, as the step $t$ increases from $0$ to the final $T_{ada}$. Each subfigure corresponds to one dataset, and different lines in one subfigure represent different training/testing splits. In the vast majority of lines, EICE decreases as $t$ increases with very few exceptions, indicating that the ensemble steps in EMQW work correctly as we expect and let the model gradually learn better $\mathbb{P}(\mathbf{y}|\mathbf{X}=x)$.

\textbf{Densities given by EMQW}. As we have always claimed, the conditional distribution given by EMQW will achieve adaptive flexibility. We plot in Fig. \ref{fig:density} the densities implied by the conditional quantiles predicted by EMQW to examine this (the density is the derivative of the inverse of quantile function, so easy to obtain numerically).
One can see that EMQW can learn distributions with various shapes: exactly Gaussian, approximately Gaussian, sharp peak, asymmetry, long tail, finite bound, multi-modality, etc. This variety may even appear within one dataset, showing the complexity of real data and the ability of EMQW.

\textbf{Densities given by other methods}. In contrast, we have claimed that many quantile-based methods with over-flexibility will give chaotic or invalid densities. We plot in Fig. \ref{fig:other_density} the densities implied by the quantiles predicted by the four quantile-based methods to verify this. Variance networks and the two Bayesian (Dropout) methods all produce Gaussian densities, so unnecessary to plot. We select one data point from the testing set of each of the three datasets: Naval, Consumption, and Air. Its densities given by the four methods are shown in Fig. \ref{fig:other_density}. From the figure, we can see they all produce rough or even invalid densities. This phenomenon is common, not just appear in these datasets.


\section{Discussion on Asymptotic Guarantee}

It is appealing to know for sure that the conditional quantiles (or conditional densities implied) approximate the true ones asymptotically. Actually, we are looking for a function $f(\cdot)=[f_1(\cdot),\cdots,f_K(\cdot)]^\top$ to minimize: 
$\min_{f\in\mathcal{F}} \mathbb{E}\left[ L(\mathbf{y},f(\mathbf{X})) \right] = \sum_{k=1}^K\mathbb{E}\left[  L_{\tau_k}(\mathbf{y},f_k(\mathbf{X})) \right]$.
Obviously, if we consider a sufficiently rich $\mathcal{F}$, the optimal $f_k(x)$ should be the $\tau_k$-quantile of $\mathbb{P}(\mathbf{y}|\mathbf{X}=x)$. In our method, $\mathcal{F}$ consists of those functions defined via a complex ensemble of $T+1$ neural networks. If we denote the actual $\tau_k$-quantile of $\mathbb{P}(\mathbf{y}|\mathbf{X}=x)$ as $q_k(x)$ and further denote $Q(x)=[q_1(x),\cdots,q_K(x)]^\top$, the consistency means that the estimated $Q^{n}(\cdot)$ given by our algorithm based on the dataset $\mathcal{D}_n=\{(\mathbf{X}_i,\mathbf{y}_i)\}_{i=1}^n$ and function class $\mathcal{F}_n$ should satisfy:
$\lim_{n\to\infty}\mathbb{E}\left[\int_{\mathbb{R}^d} \left\| Q^{n}(x)-Q(x) \right\|^2 \mu(dx)\right] = 0$,
for all distributions of $(\mathbf{X},\mathbf{y})$ with $\mathbb{E}[\mathbf{y}^2]<\infty$. Here $\mu$ denotes the distribution of $\mathbf{X}$. This is called the weak universal consistency of $Q^{n}$ (see \cite{gyorfi2002distribution} for the strong version). We assume $\mathcal{F}_n$ consists of those functions having $T_n$ ensemble steps, and $T_n\to\infty$ as $n\to\infty$.

The multi-quantile regression is not new in the literature. For example, \cite{zou2008composite,bondell2010noncrossing,liu2011simultaneous,moon2021learning} all adopted similar objectives for this. However, they share some common weaknesses: i) the function class $\mathcal{F}$ they assume is linear, linear spline, or kernel;
ii) they all add explicit monotonic constraints on quantiles in the optimization;
and iii) they all have no asymptotic analysis for the non-parametric estimators.
Our method overcomes the first two weaknesses. For the third one, \cite{chaudhuri1991nonparametric,white1992nonparametric} established the consistency of polynomial and neural-network estimators of conditional quantile, but with only one quantile level.
Here we face two challenges: i) the multi-quantile regression with many levels; and ii) the function class we consider is the ensemble of many (small) neural networks, which is a novel and unique design of us.

There have been some works analyzing the asymptotic properties of neural networks (NNs). \cite{gyorfi2002distribution} gave the consistency results of infinitely wide NNs in ordinary regression problem with least square loss. Very recently, \cite{radhakrishnan2023wide} showed that wide and deep NNs achieve consistency for classification and excessively deep NNs are harmful for regression.
\cite{lin2022universal} analyzed the consistency of deep convolutional NNs. Back to our problem, to prove the consistency of 
our estimator, we may need finishing the following three steps: 
	i) decompose $\int_{\mathbb{R}^d} \left\| Q^{n}(x)-Q(x) \right\|^2 \mu(dx)$ into the approximation error and estimation error;
	ii) prove that our function class is dense, or, our ensemble of NNs can approximate any continuous function with a compact domain;
	and iii) deal with the bound on the covering numbers of our function class.
Finishing each step will not be trivial and straightforward, especially, the third step will be challenging. Again it is because the function class we consider is the ensemble of many (small) NNs, which is quite different from the infinitely wide NN used in many theoretical works. Given these facts, we leave the asymptotic guarantee analysis to future work.


\section{Conclusions}

To conclude, we propose an ensemble multi-quantiles approach that can do adaptively flexible distribution prediction for uncertainty quantification in machine learning. It is motivated by the observations that: i) Gaussian assumption for the conditional distribution lacks flexibility; and ii) over-flexibility may be harmful for generalization, such as the invalid densities obtained by the quantile regression-based methods. We design a boosting framework for multi-quantiles prediction and carefully design the model details in the initial step and the ensemble steps. Our method is also motivated by the demand for a built-in solution to the quantile-crossing issue.

The whole ensemble approach is intuitive and interpretable, in which in the initial step a relatively strong learner produces a Gaussian distribution, and in the ensemble steps $T_{ada}$ weak learners adaptively seek a balance between the distribution structure and the flexibility. We verify these characteristics by experiments, and compare it to recent popular competing methods. Conclusively, the EMQW approach achieves start-of-the-art performance among all, under the metrics of calibration, sharpness, and tail-side calibration. Traditional ensemble tree models are also compared.

Through visualization, we find that our approach can learn distributions with various density shapes including Gaussian, approximately Gaussian, sharp peak, asymmetry, long tail, finite bound, multi-modality, etc. This variety may appear even in one dataset, confirming the complexity of real data and the necessity and merits of such an adaptively flexible ensemble approach.



%

\ifCLASSOPTIONcaptionsoff
  \newpage
\fi

\bibliographystyle{IEEEtran}
\bibliography{reference}

\end{document}